\documentclass[10pt,twocolumn,letterpaper]{article}

\usepackage{iccv}
\usepackage{times}
\usepackage{epsfig}
\usepackage{graphicx}
\usepackage{amsmath}
\usepackage{amssymb}
\usepackage{xcolor}

\usepackage{bm}
\usepackage{mathtools}
\usepackage{caption}
\usepackage{subcaption}
\usepackage{enumitem}
\usepackage{nicefrac}
\usepackage{textpos}

\usepackage{titling}
\predate{}
\postdate{}

\usepackage[pagebackref=true,breaklinks=true,letterpaper=true,colorlinks,bookmarks=false]{hyperref}

\definecolor{XL_color}{rgb}{0.858, 0.188, 0.478}
\definecolor{DM_color}{rgb}{0.9, 0.1, 0.1}

\newcommand{\T}{\bm{T}}
\newcommand{\TBA}{\prescript{B}{A}\T}
\newcommand{\TAR}{\prescript{A}{R}\T}

\newcommand{\R}{\bm{R}}
\newcommand{\RBA}{\prescript{B}{A}\R}

\newcommand{\qv}{\bm{q}}
\newcommand{\qvA}{\prescript{A}{}\qv}
\newcommand{\qvB}{\prescript{B}{}\qv}
\newcommand{\qvR}{\prescript{R}{}\qv}

\newcommand{\pv}{\bm{p}}

\newcommand{\pvB}{\prescript{B}{}\pv}

\newcommand{\cvdot}{\dot{\bm{c}}}
\newcommand{\mvdot}{\dot{\bm{m}}}
\newcommand{\mvdotA}{\prescript{A}{}\mvdot}

\newcommand{\rv}{\bm{r}}

\newcommand{\rvhat}{\hat{\rv}}

\newcommand{\rdot}{\dot{r}}

\newcommand{\dt}{\Delta t}

\newcommand{\Paragraph}[1]{\vspace{1mm} \noindent \textbf{#1} \hspace{0mm}}
\newcommand{\Section}[1]{\vspace{-1mm} \section{#1} \vspace{0mm}}
\newcommand{\SubSection}[1]{\vspace{-1mm} \subsection{#1} \vspace{-0mm}}
\newcommand{\SubSubSection}[1]{\vspace{-1mm} \subsubsection{#1} \vspace{-1mm}}

\newcommand\Mark[1]{\textsuperscript#1}

\iccvfinalcopy 

\def\iccvPaperID{10568} 

\begin{document}
	
\begin{textblock*}{\textwidth}(0cm,0cm)
	\large\noindent{\copyright~2021 IEEE.  Personal use of this material is permitted.  Permission from IEEE must be obtained for all other uses, in any current or future media, including reprinting/republishing this material for advertising or promotional purposes, creating new collective works, for resale or redistribution to servers or lists, or reuse of any copyrighted component of this work in other works.}
\end{textblock*}
\thispagestyle{empty}

\title{\textbf{Full-Velocity Radar Returns by Radar-Camera Fusion}}

\author{
Yunfei Long\Mark{1}, Daniel Morris\Mark{1},  Xiaoming Liu\Mark{1}, \\
Marcos Castro\Mark{2},
Punarjay Chakravarty\Mark{2},
and Praveen Narayanan\Mark{2} \\ 
\Mark{1}Michigan State University, \Mark{2}Ford Motor Company \\
{\tt\small \{longyunf,dmorris,liuxm\}@msu.edu},
{\tt\small \{mgerard8,pchakra5,pnaray11\}@ford.com}
}
\date{}

\maketitle
\pagenumbering{arabic}

\begin{abstract}
A distinctive feature of Doppler radar is the measurement of velocity in the radial direction for radar points. 
However, the missing tangential velocity component hampers object velocity estimation as well as temporal integration of radar sweeps in dynamic scenes.  
Recognizing that fusing camera with radar provides complementary information to radar, in this paper we present a closed-form solution for the point-wise, full-velocity estimate of Doppler returns using the corresponding optical flow from camera images.  
Additionally, we address the association problem between radar returns and camera images with a neural network that is trained to estimate radar-camera correspondences. 
Experimental results on the nuScenes dataset verify the validity of the method and show significant improvements over the state-of-the-art in velocity estimation and accumulation of radar points.   
\end{abstract}

\section{Introduction}
Radar is a mainstream automotive 3D sensor, and along with LiDAR and camera, is used in perception systems for driving assistance and autonomous driving~\cite{stanislas2015characterisation, li2020lidar,m3d-rpn-monocular-3d-region-proposal-network-for-object-detection}. 
Unlike LiDAR, radar has been widely installed on existing vehicles due to its relatively low cost and small sensor size, which makes it an easy fit into various vehicles without changing their appearance. Thus, advances in radar vision systems have potential to make immediate impact on vehicle safety. Recently, with the release of a couple of autonomous driving datasets with radar data included,~\emph{e.g.}, Oxford Radar RobotCar~\cite{barnes2020oxford} and nuScenes~\cite{caesar2020nuscenes}, there is great interest in the community to explore how to leverage radar data in various vision tasks such as object detection~\cite{nabati2021centerfusion, yang2020radarnet}.

\begin{figure}[t!]
    \captionsetup{font=small}
	\centering
	\begin{subfigure}[b]{\linewidth}
		\centering
		\includegraphics[width=0.9\textwidth]{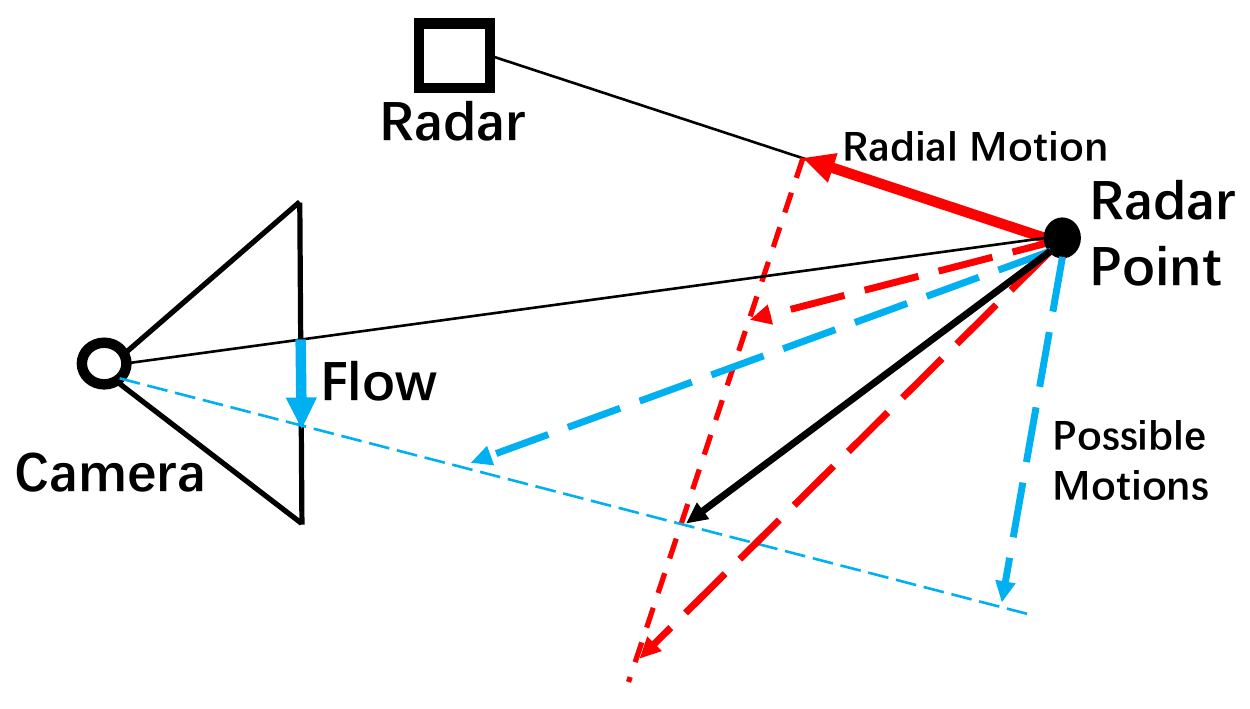}
		\vspace{-2mm}
		\caption{$ $}
		\label{fig:y equals x}
	\end{subfigure}
	\hfill
	\begin{subfigure}[b]{0.2\textwidth}
		\centering
		\includegraphics[width=\textwidth]{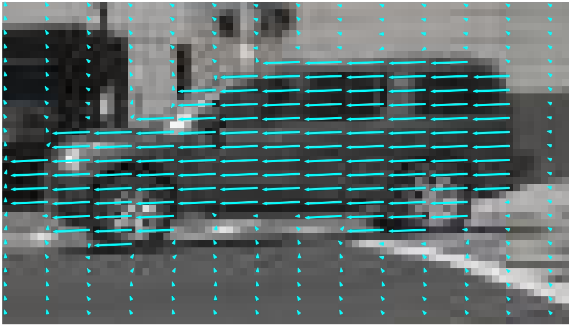}
		\caption{$ $}
		\label{fig:three sin x}
	\end{subfigure}
	\hfill
	\begin{subfigure}[b]{0.22\textwidth}
		\centering
		\includegraphics[width=\textwidth]{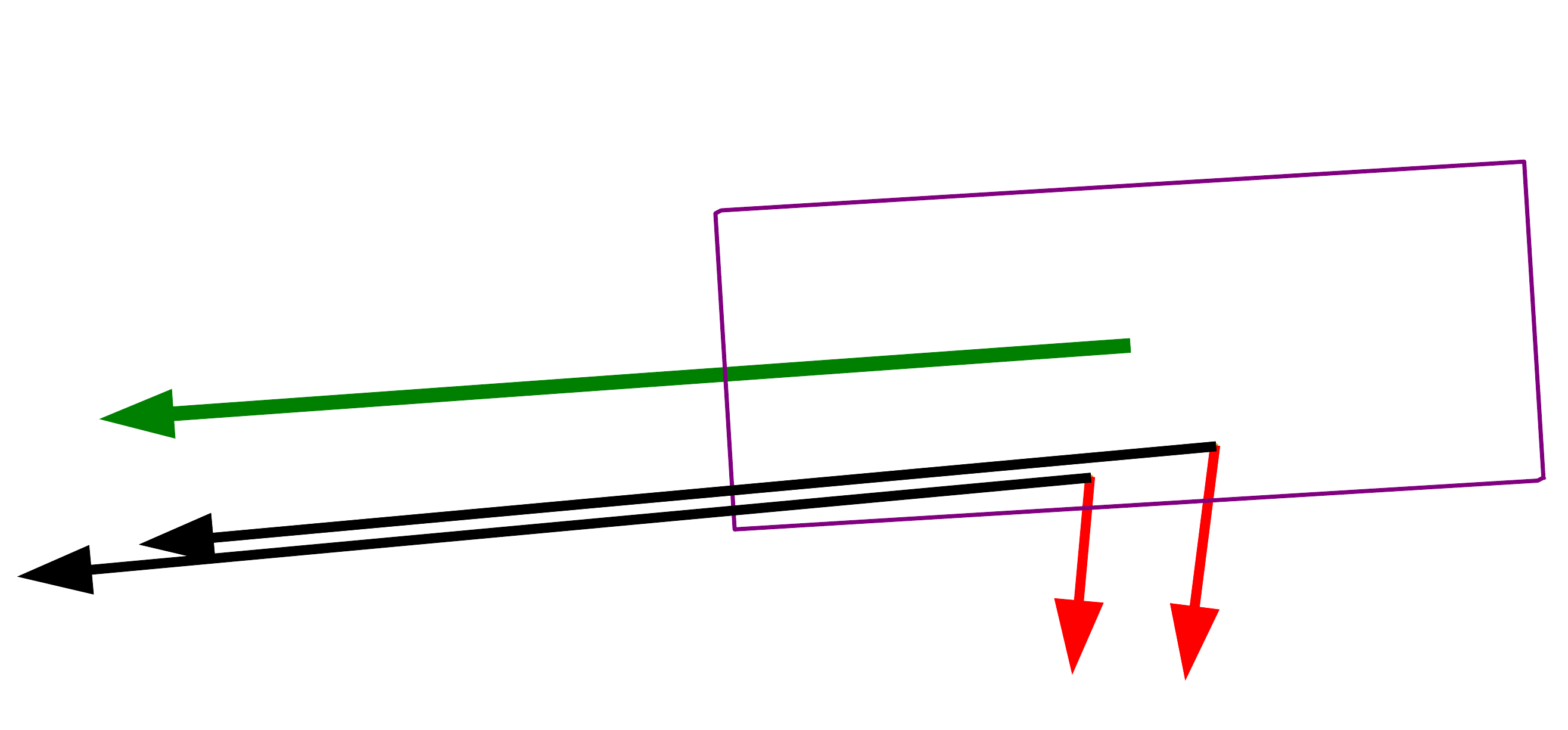}
		\caption{$ $}
		\label{fig:five over x}
	\end{subfigure}
	\vspace{-2mm}
	\caption{\small (a) Full motion cannot be determined with a single sensor: all motions ending on the blue dashed line (\emph{i.e.},~blue dashed arrows) map to the same optical flow and all motions terminated on the red dashed line (\emph{i.e.},~red dashed arrows) fit the same radial motion. However, with a radar-camera pair, the full motion can be uniquely decided: only the motion drawn in black satisfies both optical flow and radial motion. (b) Optical flow in the camera-image and (c) a bird's-eye view of the observed vehicle. This shows measured radar points with radial velocity (red), our predicted point-wise, full velocity (black), and ground truth full velocity of the vehicle (green).}
	\vspace{-3mm}
	\label{Figure:indeterminacy}
\end{figure}

In addition to measuring 3D positions, radar has the special capability of obtaining radial velocity of returned points based on the Doppler effect. 
This extra capability is a significant advantage over other 3D sensors like LiDAR, enabling, for instance, instantaneous moving object detection. 
However, due to the inherently ambiguous mapping from radial velocity to full velocity, using radial velocity directly to account for the real movement of radar points is inadequate and sometimes misleading. 
Here, the full velocity denotes the actual velocity of radar points in 2D or 3D space. 
While radial velocity can well approximate full velocity when a point is moving away from or towards the radar, these two can be very different when the point is moving in the non-radial directions. 
An extreme case occurs for objects moving tangentially as these will have zero radial velocity regardless of target speed. Therefore, acquiring point-wise full velocity instead of radial velocity is crucial to reliably sense the motion of surrounding objects.

Apart from measuring the velocity of objects, another important application of point-wise velocity is the accumulation of radar points. 
Radar returns from a single frame are much sparser than LiDAR in both azimuth and elevation,~\emph{e.g.}, typically LiDAR has an azimuth resolution $10\times$ higher than radar~\cite{yang2020radarnet}.
Thus, it is often essential to accumulate multiple prior radar frames to acquire sufficiently dense point clouds for downstream tasks,~\emph{e.g.}, object detection~\cite{nobis2019deep, chadwick2019distant, chang2020spatial}. 
To align radar frames, in addition to compensating egomotion, we shall consider the motion of moving points in consecutive frames, which can be estimated by point-wise velocity and time of movement. 
As the radial velocity does not reflect the true motion, it is desirable to have point-wise full velocity for point accumulation.

To solve the aforementioned dilemma of radial velocity, we propose to estimate point-wise full velocity of radar returns by fusing radar with a RGB camera. 
Specifically, we derive a closed-form solution to infer point-wise full velocity from radial velocity as well as associated projected image motion obtained from optical flow. As shown in Fig.~\ref{Figure:indeterminacy}, constraints imposed by optical flow resolve the ambiguities of radial-full velocity mapping and lead to a unique and closed-form solution for full velocity. 
Our method can be considered as a way to enhance raw radar measurement by upgrading point-wise radial velocity to full velocity, laying the groundwork for improving radar-related tasks, ~\emph{e.g.}, velocity estimation, point accumulation and object detection. 

Moreover, a prerequisite for our closed-form solution is the association between moving radar points and image pixels. 
To enable a reliable association, we train a neural network to predict radar-camera correspondences as well as discerning occluded radar points. 
Experimental results demonstrate that the proposed method improves point-wise velocity estimates and their use for object velocity estimation and radar point accumulation.

In summary, the main contributions of this work are:
\begin{itemize} 
    \item We define a novel research task for radar-camera perception systems,~\emph{i.e.}, estimating point-wise full velocity of radar returns by fusing radar and camera.
    
    \item We propose a novel closed-form solution to infer full radar-return velocity by leveraging the radial velocity of radar points, optical flow of images, and the learned association between radar points and image pixels.
    
    \item We demonstrate state-of-the-art (SoTA) performance in object velocity estimation, radar point accumulation, and 3D object localization.
    
\end{itemize}

\section{Related Works}

\Paragraph{Application of Radar in Vision}
Radar data differs from LiDAR data in various aspects~\cite{brodeski2019deep}. 
In addition to the popular point representation (also named radar target~\cite{palffy2020cnn}), an analogy to LiDAR points, there are other radar data representations containing more raw measurements, {\it e.g.}, range-azimuth image and spectrograms, which have been applied in tasks such as activity classification~\cite{seyfiouglu2018deep}, detection~\cite{lim2019radar}, and pose estimation~\cite{roos2016reliable}.  
Our method is based on radar points, with the format available in the  nuScenes dataset~\cite{caesar2020nuscenes}. 

The characteristics of radar have been explored to complement other sensors. 
The Doppler velocity of radar points is used to distinguish moving targets. 
For example, RSS-Net~\cite{kaul2020rss} uses radial velocity as a motion cue for image semantic segmentation. Chadwick~\emph{et al.}~\cite{chadwick2019distant} use radial velocity to detect distant moving vehicles---difficult to detect with only images. 
Fritsche~\emph{et al.}~\cite{fritsche2017fusion} combine radar with LiDAR for measurement under poor visibility. 
With a longer detection range than LiDAR, radar is also deployed with LiDAR to better detect far objects~\cite{yang2020radarnet}.

The {\it sparsity} of radar makes it difficult to directly apply well-developed techniques for LiDAR on radar~\cite{lim2019radar, nabati2021centerfusion}. 
For example, Danzer~\emph{et al.}~\cite{danzer20192d} adopt PointNets~\cite{qi2017pointnet} on radar points for 2D car detection, while sparsity limits it to large objects like cars. 
Similar to LiDAR-camera depth completion~\cite{depth-completion-with-twin-surface-extrapolation-at-occlusion-boundaries,depth-coefficients-for-depth-completion}, Long~\emph{et al.}~\cite{long2021radar} develop radar-camera depth completion by learning a probabilistic mapping from radar returns to images.
To obtain denser radar points, Lombacher~\emph{et al.}~\cite{lombacher2016potential} use occupancy grid~\cite{elfes1989using} to accumulate radar frames.
Yet, the method assumes a static scene and cannot cope with moving objects. 
Radar points are projected on images and represented as regions near projected points, such as vertical bars~\cite{nobis2019deep} and circles~\cite{chadwick2019distant, chang2020spatial}, to account for uncertainty of projection due to measurement error. 
While accumulating radar frames is desirable, without reliably compensating object motion, these methods need to carefully decide the number of frames to trade off between the gain in accumulation and loss in accuracy due to delay~\cite{nobis2019deep}. 
Our estimated point-wise velocity can compensate object motion and realize more accurate accumulation.

\Paragraph{Velocity Estimation in Perception Systems}
Researchers have used monocular videos~\cite{kinematic-3d-object-detection-in-monocular-video} or radial velocity of radar points to estimate {\it object-wise} velocity. With only radar data of a single frame, Kellner~\emph{et al.}~\cite{kellner2013instantaneous, kellner2014instantaneous} compute full velocity of moving vehicles from radial velocities and azimuth angles of at least two radar hits. 
However, for a robust solution, the method requires that 1) radar captures more radar hits on each object, 2) radar points have significantly different azimuth angles and 3) object points are clustered before velocity estimation~\cite{kellner2013instantaneous, schlichenmaier2019clustering, scheiner2019multi}. 
Obviously due to sparsity of radar in a single frame, it is difficult to obtain at least two radar hits on distant vehicles, let alone objects of smaller sizes. 
Also, it is common that radar points on the same object, {\it e.g.}, a distant or small object, have similar azimuth. 

\begin{figure*}[t!]
    \captionsetup{font=small}
	\begin{center}
		\includegraphics[width=\linewidth]{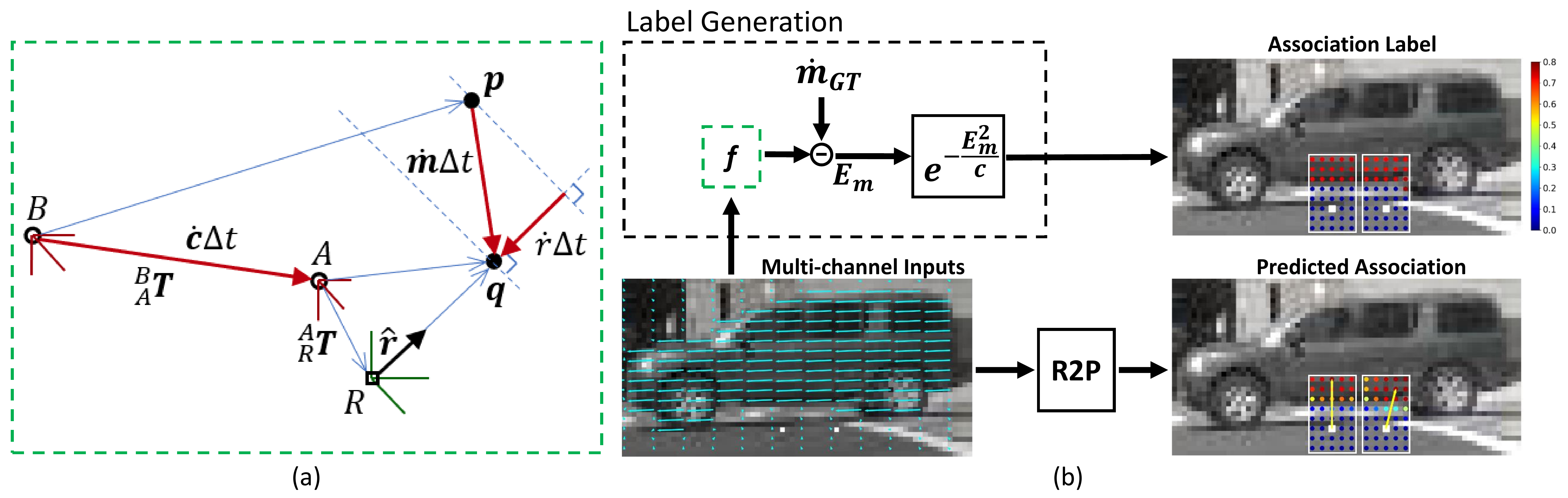}
	\end{center}
	\vspace{-6mm}
	\caption{\small \textbf{Full velocity estimation and learning to associate radar points to camera pixels}. (a) A 3D point, $\pv$, is observed by a camera at $B$.  A short interval, $\Delta t$, later, the point has moved by $\mvdot\dt$ to $\qv$ while the camera has moved by $\cvdot\dt$ to $A$. At the same time, the radar measures both the position of $\qv$ and the radial speed $\rdot$, which is the radial component of $\mvdot$. Using radial speed $\rdot$ and the associated  optical flow of $\qv$ in images, we derive a closed-form equation (denoted as $\bm{f}()$) to estimate $\qv$'s full velocity $\mvdot$. (b) As the closed-form solution requires point-wise association of two sensors, we train a Radar-2-Pixel (R2P) network to take a multi-channel input and predict the association probabilities for pixels within a neighborhood of the raw projection (white dot) obtained via known pose $\TAR$. A pixel with the highest probability (yellow arrow) is deemed as the associated pixel of a radar point. To obtain labels for training R2P, our label generation module uses $\bm{f}()$ to compute velocities of all neighboring pixels, then calculates velocity error $E_m$ by using the ground truth velocity $\mvdot_{GT}$, and finally obtains association probabilities of these neighbors based on $E_m$.}
	\label{fig:diagram_association}
		\vspace{-3mm}
\end{figure*}

Recognizing the density and accuracy limitation of radar, researchers fuse radar with other sensors, {\it e.g.}, LiDAR and camera, for object-wise velocity estimation. 
Specifically, existing techniques~\cite{zhao2019object,wu2020deep,li2020deep} for images or LiDAR are employed to obtain preliminary detections. 
Radar data, including radial velocity, once associated with the initial detections, are used as additional cues to predict full velocities of objects. 
For instance, in RadarNet~\cite{yang2020radarnet} 
temporal point clouds of radar and LiDAR, modeled as voxels, are used to acquire initial detections and their motions. 
Object motion direction is used to resolve the ambiguities in radar-point association by
back-projecting their radial velocities on the motion direction. 
Yet, a sequence of LiDAR frames is required to obtain the initial detection and motion estimation.

CenterFusion~\cite{nabati2021centerfusion} integrates radar with camera for object-wise velocity estimation. 
Well-developed image-based detector is applied to extract preliminary boxes.
After associating radar points with detections, the method combines radar data, radial velocity and depth, with image features within detected regions to regress a full velocity per detection. 
However, without a closed-form solution, the mapping from radial to full velocity needs to be learned from a great number of labeled data. 
In contrast, we present a point-wise {\it closed-form solution} for full-velocity estimation of radar points, without performing object detection. 
To our knowledge, there is no prior method able to perform point-wise full-velocity estimation for radar returns.

\section{Proposed Method}

We consider the case of a camera and radar rigidly attached to a moving platform, {\it e.g.}, a vehicle, observing moving objects in the environment.  In this section we develop equations relating optical flow measurements in the camera to position and velocity measurements made by the radar.  

\subsection{Physical Configuration and Notation}

The physical configuration of our camera and radar measurements is illustrated in Fig.~\ref{fig:diagram_association}(a). Three coordinate systems are shown: $A$ and $B$ specifying camera poses and $R$ specifying a radar pose.  The camera at $B$ observes a 3D point $\pv$.  A short interval later, $\dt$, the point has moved to $\qv$, the camera to $A$ and the radar to $R$, and both the camera and radar observe the target point $\qv$.  These 3D points are specified by $4$-dim homogeneous vectors, and when needed, a left-superscript specifies the coordinate system in which it is specified, {\it e.g.}, $\prescript{A}{}\qv$ indicates a point relative to a coordinate system $A$.  The target velocity, $\mvdot$, and camera velocity $\cvdot$ are specified by $3$-dim vectors, again optionally with a left superscript to specify a coordinate system.    

Coordinate transformations, containing both a rotation and translation, are specified by $4\times 4$ matrices, such as $\prescript{B}{A}{\T}$, which transforms points from the left-subscript coordinate system to the left-superscript coordinate system.  In this case we transform a point from $A$ to $B$ with:
\begin{equation}
    \prescript{B}{}\qv = \prescript{B}{A}{\T} \: \prescript{A}{}\qv.
\end{equation}
Only the rotational component of these transformations is needed to transform velocities.  For example, $\prescript{A}{}\mvdot$ is transformed to $\prescript{B}{}\mvdot$ by the $3\times 3$ rotation matrix $\prescript{B}{A}{\R}$:
\begin{equation}
    \prescript{B}{}\mvdot = \prescript{B}{A}{\R} \: \prescript{A}{}\mvdot.
\end{equation}

A vector with a right subscript, {\it e.g.}, ${\pv_i}$, indicates the $i$'th element of ${\pv}$, while a right subscript of ``1:3'' puts the first $3$ elements in a $3$-dim vector.  
For a matrix, the right subscript indicates the row. Thus $\prescript{B}{A}{\R_i}$ is a $1\times 3$ row vector containing its $i$-th row.  A right superscript ``${}^\mathsf{T}$'' is a matrix transpose.

The projections of points $\pv$ and $\qv$ are specified in either undistorted raw pixel coordinates, {\it e.g.}, $(x_q,y_q)$ or their normalized image coordinates $(u_q,v_q)$ given by:
\begin{equation}
    u_q = (x_q-c_x) / f_x, \hspace{0.5cm} v_q = (y_q-c_y) / f_y.
\end{equation}
Here $c_x,c_y,f_x,f_y$ are intrinsic camera parameters, while the right subscript of the pixel refers to the point being projected.  Vectors for 3D points can be expressed in terms of the normalized image coordinates:
\begin{align}
\qvA = \begin{pmatrix} u_qd_q\\ v_qd_q\\ d_q\\ 1 \end{pmatrix} 
\;\; \text{and} \;\;
\pvB = \begin{pmatrix} u_pd_p\\ v_pd_p\\ d_p\\ 1 \end{pmatrix}.
\label{eq:qa}
\end{align}
Here $d_q$ and $d_p$ are depths of points $\qvA$ and $\pvB$ respectively. 

We assume dense optical flow is available that maps target pixel coordinates observed in $A$ to $B$ as follows:
\begin{equation}
    \text{Flow}\left( (u_q,v_q) \right) \rightarrow (u_p,v_p).
    \label{eq:flow}
\end{equation}
Further, we assume the following are known: camera motion, $\TBA$, relative radar pose, $\TAR$, and intrinsic parameters.

\subsection{Full-Velocity Radar Returns}

The Doppler velocity measured by a radar is just one component of the three-component, full-velocity vector of an object point.  Here our goal is to leverage optical flow from a synchronized camera to augment radar and estimate this full-velocity vector for each radar return.  

\subsubsection{Relationship of Full Velocity to Radial Velocity}

The target motion from $\pv$ to $\qv$ is modeled as constant velocity, $\mvdot$, over time $\dt$, such that 
\begin{equation}
    \mvdot=\frac{\qv_{1:3}-\pv_{1:3}}{\dt}.  
    \label{eq:velocity}
\end{equation}
Our goal is to estimate the full target velocity, $\mvdot$.  Radar provides an estimate of the target position, $\qv$, but not the previous target location $\pv$.  Radar also provides the signed radial speed, $\rdot$, which is one component of $\mvdot$.  In the nuScenes dataset $\rdot$ is given by:
\begin{equation}
    \rdot = \rvhat^\mathsf{T}\mvdot.
    \label{eq:radial}
\end{equation}
Here $\rvhat$ is the unit-norm vector along the direction to the target $\qvR$.   Note that this equation is coordinate-invariant, and could  be equally written in $A$ using $\prescript{A}{}\rvhat$ and $\mvdotA$.  Now Eq.~\eqref{eq:radial} is actually the egomotion-corrected Doppler speed.  The raw Doppler speed, $\rdot_{raw}$, is the radial component of the \emph{relative} velocity between target and sensor, $\mvdot-\cvdot$, and this constraint is given by:
\begin{equation}
    \rdot_{raw} = \rvhat^\mathsf{T}(\mvdot-\cvdot),
    \label{eq:radial_relative}
\end{equation}
where $\cvdot$ is the known ego-velocity. Either Eq.~\eqref{eq:radial} or \eqref{eq:radial_relative} can be used in our formulation, depending on whether $\rdot$ or $\rdot_{raw}$ is available from the radar.

\subsubsection{Relationship of Full Velocity to Optical Flow}

In solving the velocity constraints, we first identify the known variables.  The radar measures $\qvR$, and transforming this we obtain $\qvA=\prescript{A}{R}\T\:\qvR$ which contains $d_q$ as the third component.  Image coordinates $(u_q,v_q)$ are obtained by projection, and using optical flow in Eq.~\eqref{eq:flow}, we can also obtain the $(u_p,v_p)$ components of $\pvB$.  The key parameter we do not know from this is the depth, $d_p$, in $B$.

Next we eliminate this unknown depth from our constraints. Eq.~\eqref{eq:velocity} can be rearranged and each component expressed in frame $B$:
\begin{equation}
    \pvB_{1:3} = \qvB_{1:3} - \RBA\: \mvdotA\dt,
    \label{eq:pvB2}
\end{equation}
where the second term on the right is the transformation of the target motion into $B$ coordinates.  The third row of this equation is an expression for $d_p$:
\begin{equation}
    d_p = \qvB_3 - \RBA_3\: \mvdotA\dt.
    \label{eq:dp1}
\end{equation}
Substituting this for $d_p$, and the components of $\pvB$ from  Eq.~\eqref{eq:qa}, into the first two rows of Eq.~\eqref{eq:pvB2}, we obtain
\begin{eqnarray}
\begin{bmatrix}
u_p(\qvB_3 - \RBA_3\: \mvdotA\dt) \\
v_p(\qvB_3 - \RBA_3\: \mvdotA\dt) \\
\end{bmatrix}
=
\begin{bmatrix}
\qvB_1 - \RBA_1\: \mvdotA\dt \\
\qvB_2 - \RBA_2\: \mvdotA\dt \\
\end{bmatrix},
\label{eq:dp2}
\end{eqnarray}
and rearrange to give two constraints on the full velocity:
\begin{eqnarray}
\begin{bmatrix}
\RBA_1 - u_p\RBA_3 \\
\RBA_2 - v_p\RBA_3 \\
\end{bmatrix}
\mvdotA
=
\begin{bmatrix}
\left(\qvB_1 - u_p\qvB_3\right)/\dt  \\
\left(\qvB_2 - v_p\qvB_3\right)/\dt \\
\end{bmatrix}.
\label{eq:dp3}
\end{eqnarray}

\subsubsection{Full-Velocity Solution}

We obtain three constraints on the full velocity, $\mvdotA$, from Eq.~\eqref{eq:dp3} and by converting Eq.~\eqref{eq:radial}  to $A$ coordinates. Combining these we obtain: 
\begin{eqnarray}
\begin{bmatrix}
\RBA_1 - u_p\RBA_3 \\
\RBA_2 - v_p\RBA_3 \\
\prescript{A}{}\rvhat^\mathsf{T} \\
\end{bmatrix}
\mvdotA
=
\begin{bmatrix}
\left(\qvB_1 - u_p\qvB_3\right)/\dt  \\
\left(\qvB_2 - v_p\qvB_3\right)/\dt \\
\rdot \\
\end{bmatrix}.
\label{eq:dp4}
\end{eqnarray}
Then inverting the $3\times 3$ coefficient of $\mvdotA$ gives a closed form solution for the full velocity:
\begin{eqnarray}
\mvdotA =
\begin{bmatrix}
\prescript{B}{A}{\bm{R}_1} - u_p \prescript{B}{A}{\bm{R}_3}\\
\prescript{B}{A}{\bm{R}_2} - v_p \prescript{B}{A}{\bm{R}_3}\\
\prescript{A}{}\rvhat^\mathsf{T}\\ 
\end{bmatrix}^{-1}
\begin{bmatrix}
\left(\prescript{B}{}{\qv}_1 - u_p \prescript{B}{}{\qv_3} \right) / \dt\\
\left(\prescript{B}{}{\qv}_2 - v_p \prescript{B}{}{\qv_3} \right) / \dt\\
\rdot \\ 
\end{bmatrix}
\label{eq:full_v}.
\end{eqnarray}

\begin{figure}[t!]
    \captionsetup{font=small}
	\centering
	\scalebox{1}{
		\begin{tabular}{@{}c@{}c@{}c@{}c@{}}
			\includegraphics[width=1.4 in]{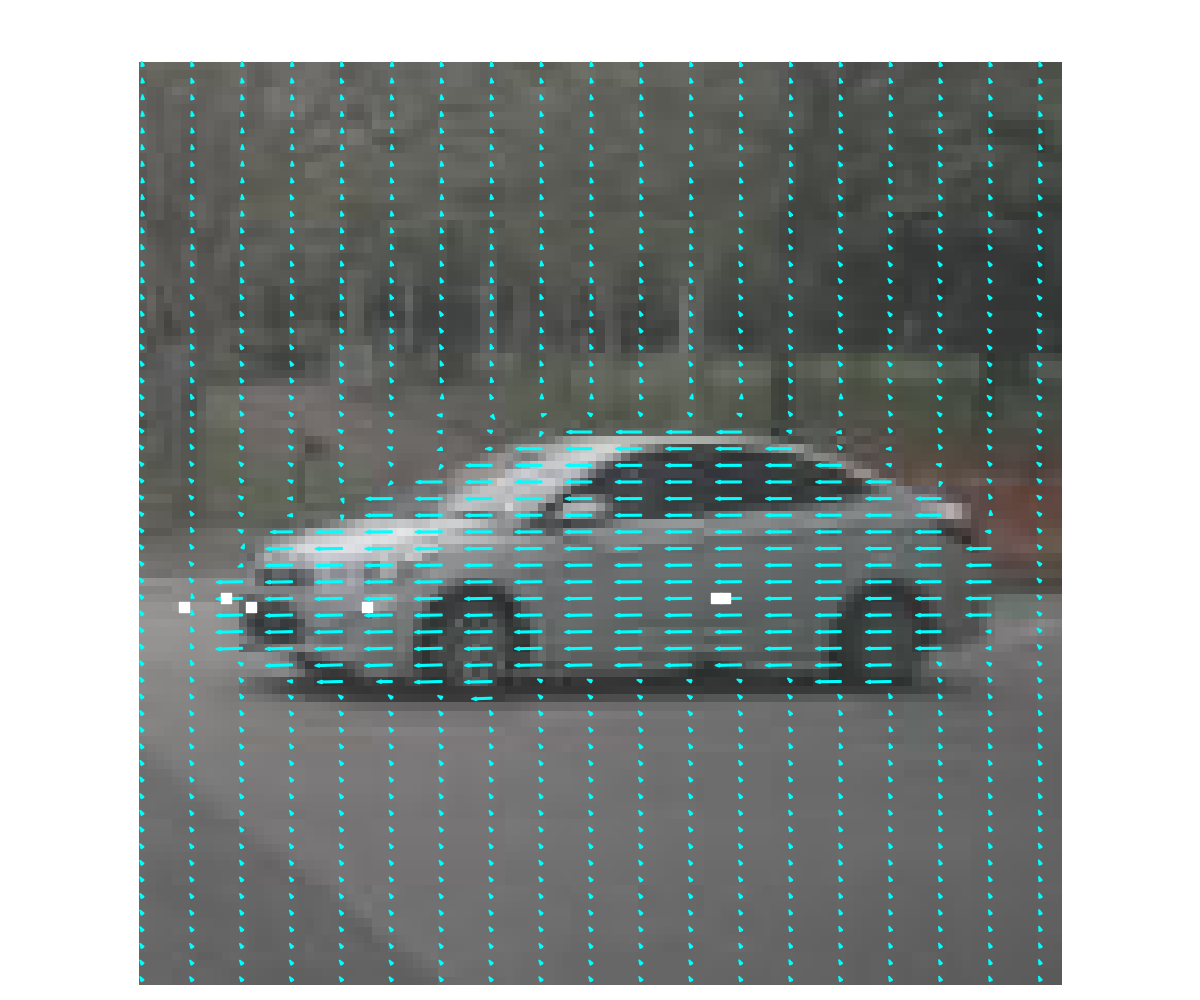}  &
			\includegraphics[width=1.4 in]{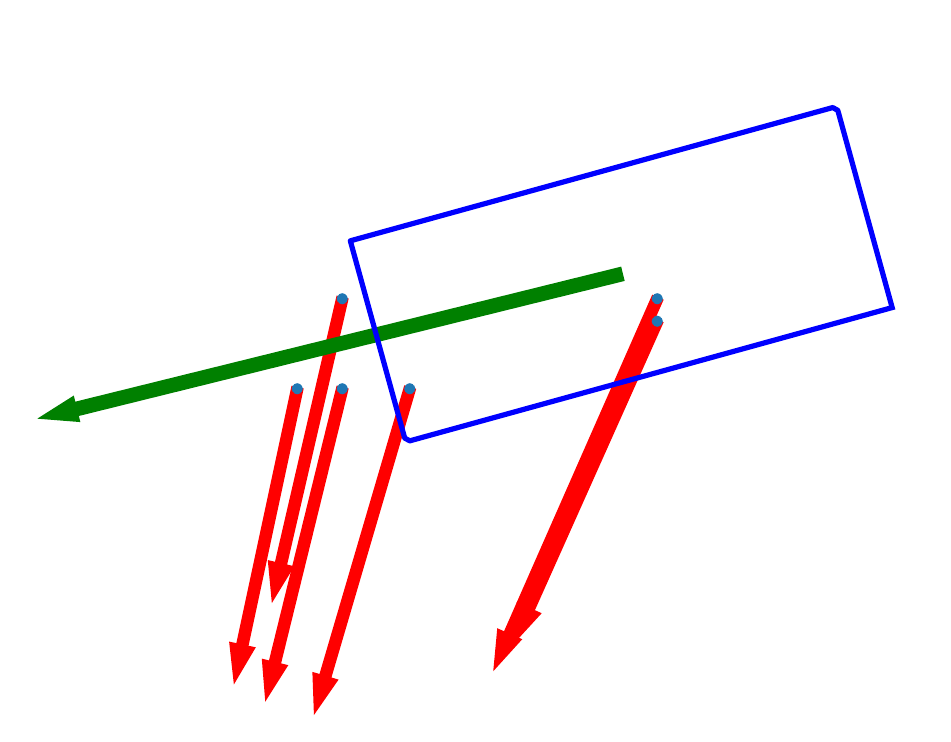} \vspace{-1mm}\\
			\footnotesize{ (a) } &  \footnotesize{ (b) } \\
			\includegraphics[width=1.4 in]{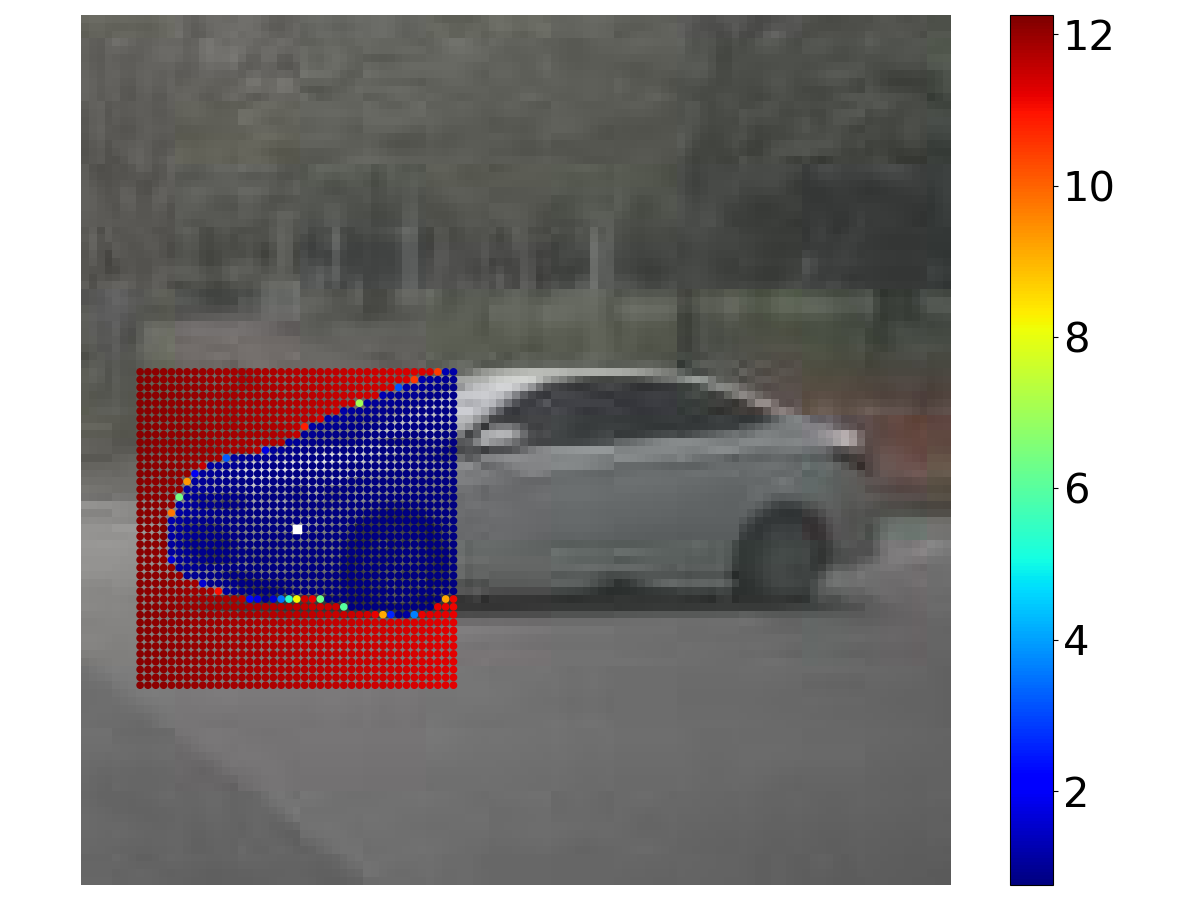}  &
			\includegraphics[width=1.4 in]{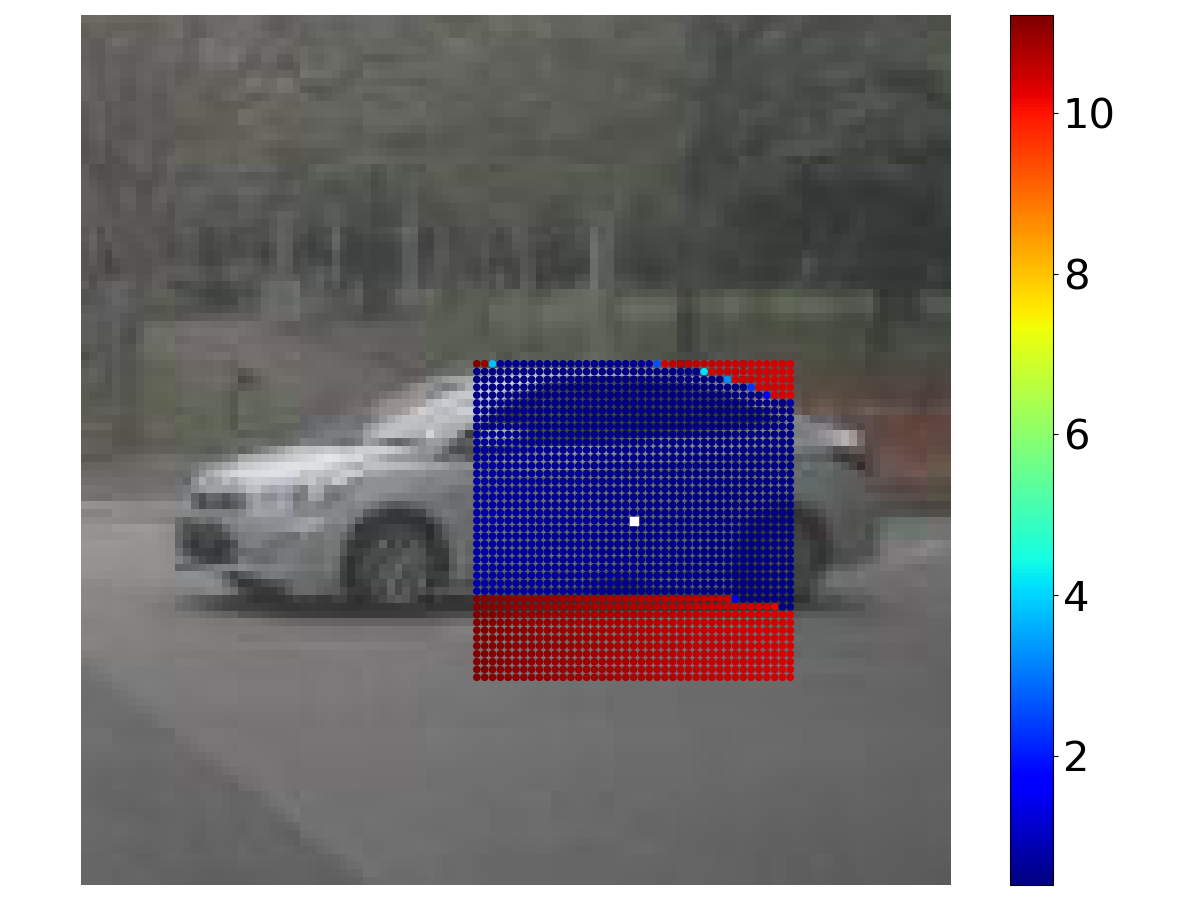} \vspace{-1mm}\\
			 \footnotesize{ (c) } & \footnotesize{ (d) } \\
	\end{tabular}   }
	\vspace{-2mm}
	\caption{\small (a) Optical flow; (b) Bird's-eye view of GT bounding box, radial velocity (red) and GT velocity (green); (c) and (d) show $E_m$, computed by using Eq.~(\ref{eq:e_v}), for two radar projections (white square) over $41\times41$ pixel regions, respectively. For radar hits reflected from the vehicle, $E_m$ is small for neighboring pixels on the car and large on the background.}
	\label{Figure:label}\vspace{-3mm}
\end{figure}

Recall in Fig.~\ref{Figure:indeterminacy}(a) the red/blue dashed lines show the velocity constraints from radar/flow.  The solution of Eq.~\eqref{eq:full_v} is the full velocity that is consistent with both constraints. 
We note that this can handle moving sensors, although Fig.~\ref{Figure:indeterminacy}(a) shows the case of a stationary camera for simplicity.  Further, if we set $\dt<0$, Eq.~\eqref{eq:full_v} also applies to the case that the point shifts from $\qv$ to $\pv$ as the camera moves from $A$ to $B$.  And one limitation is that Eq.~\eqref{eq:full_v} cannot estimate full velocity for radar points occluded in the camera view, although we can typically identify those occlusions.

\SubSection{Image Pixels and Radar Points Association}

Our solution for point-wise velocity in Eq.~(\ref{eq:full_v}) assumes that we know the pixel coordinates $(u_q,v_q)$ of the radar-detected point, $\prescript{R}{}\qv$.
It appears straightforward to obtain this pixel correspondence by projecting a radar point onto the image using the known radar-image coordinate transformation, $\prescript{A}{R}\T$. 
We refer to this corresponding pixel as ``raw projection''. 
However, there are a number of reasons why raw projection of radar points into an image is inaccurate. Radar beam-width typically subtends a few degrees and is large relative to a pixel, resulting in low resolution target location in both azimuth and elevation.  Also, a radar displaced from a camera can often see behind an object, as viewed by the camera, and when these returns are projected onto an image they incorrectly appear to correspond to the foreground occluding object. Using flow from an occluder or an incorrectly associated object pixel may result in incorrect full-velocity estimation.  To address these issues with raw projection, we train a neural network model, termed Radar-2-Pixel (R2P) network, to estimate associated radar pixels in the neighborhood of raw projection and identify occluded radar points.
Similar models have been applied to image segmentation~\cite{kampffmeyer2018connnet} and radar depth enhancement~\cite{long2021radar}.

\SubSubSection{Model Structure}
Our method estimates association probabilities (ranging from $0$ to $1$) between a moving radar point and a set of pixels in the neighborhood of its raw projection. 
The R2P network is an encoder-decoder structure with inputs and outputs of image resolution. 
Stored in $8$ channels, the input data include image, radar depth map (with depth on raw projections) and optical flow. 
The output has $N$ channels, representing predicted association probability for $N$ pixel neighbors. 
The association between the radar point, $\qvA$, and the $k$-th neighbor of raw projection $(x,y)$ is stored in $A(x,y,k)$, where $k=1,2,...,N$.

\SubSubSection{Ground Truth Velocity of Moving Radar Points}
\label{sec:GT_velocity}
The nuScenes~\cite{caesar2020nuscenes} provides the GT (ground truth) velocity of object bounding boxes. 
We associate radar hits on an object to its labeled bounding box, and assign the velocity of the box to its associated radar points.
The association is determined based on two criteria:
1) in radar coordinates, the distance between radar points and associated box is smaller than a threshold $T_d$; 
and 2) the percentage error between the radial velocity of a radar point and the radial component of the velocity of associated box is smaller than a threshold $T_p$.

\SubSubSection{Generating Association Labels}
We can project a radar point expressed in corresponding camera coordinates, $\qvA$, to pixel coordinates $(u_q,v_q)$, but as mentioned before, often this image pixel does not correspond to the radar return.  Our proposed solution is to search in a neighboring region around $(u_q,v_q)$ for a pixel whose motion is consistent with the radar return.  This neighborhood search is shown in Fig.~\ref{fig:diagram_association}.   If a pixel is found, then we correct the 3D radar location $\qvA$ to be consistent with this pixel, otherwise we mark this radar return as occluded.

We learn this radar-to-pixel association and correction by training the R2P network.  
We generate true association score between a radar point and a pixel according to the compatibility between the true velocity and the optical flow at that pixel: high compatibility indicates high association. 
To quantify the compatibility, assuming a pixel is associated with a radar point, we compute a hypothetical full velocity for the radar point by using the optical flow of that pixel according to Eq.~\eqref{eq:full_v}. 
The flow is considered compatible
if the hypothetical velocity is close to the GT velocity. 
Specifically, the  hypothetical velocity can be computed as
\begin{equation}
\resizebox{.88\hsize}{!}{
$\mvdotA_{est}(x,y,k) = \bm{f}\left(\breve{u}_q,\breve{v}_q,\breve{u}_p,\breve{v}_p, d_q, \rdot, \TBA, \TAR \right),$
}
\label{eq:fvel}
\end{equation}
where $k=1,\cdots,N$, $\bm{f}(\cdot)$ is the function to solve full velocity via Eq.~\eqref{eq:full_v}, and $(x,y)$ is the raw projection of the radar point.
Note that  $\breve{u}_q=u_q\left[x+\Delta x(k), y+\Delta y(k)\right]$, $\breve{v}_q$ is defined similarly, 
and $[\Delta x(k),\Delta y(k)]$ is the coordinate offset from raw projection to the $k$-th neighbor.  Using flow, Eq.~\eqref{eq:flow}, we obtain $(\breve{u}_p,\breve{v}_p)$ from $(\breve{u}_q,\breve{v}_q)$.

Second, we calculate the $L_2$ norm of errors between $\mvdotA_{est}(x,y,k)$ and ground truth velocity $\mvdotA_{GT}(x,y)$ by
\begin{equation}
E_m(x,y,k)= \lVert \mvdotA_{est}(x,y,k) - \mvdotA_{GT}(x,y) \rVert_2
\label{eq:e_v}.
\end{equation}
Fig.~\ref{Figure:label} shows examples of $E_m$ for two radar hits on a car.

Finally, we transform $E_m$ to an association score with
\begin{equation}
L(x,y,k)= e^{-\frac{E^2_m(x,y,k)}{c}}
\label{eq:prob},
\end{equation}
where $L$ is used as a label for association probability between a radar and its $k$-th neighbor. Note that $L$ increases with decreasing $E_v$, and $c$ is a parameter adjusting the tolerance of velocity errors when converting errors to association. 
We use the cross entropy loss to train the model.

\SubSubSection{Estimate Association and Identify Occlusion}
With a trained model, we can estimate association probability between radar points and $N$ pixels around their raw projections $(x,y)$, {\it i.e.}, $A(x,y,k)$.
Among the $N$ neighbors, the radar return velocity may be compatible with a number of pixels, and we select the pixel with the maximum association, $A_{max}$, as the neighbor ID $k_{max}$:
\begin{equation}
k_{max}= \underset{k}{\arg\max}[A(x,y,k)]
\label{eq:k_max}.
\end{equation}
 If $A_{max}$ is equal or larger than a threshold $T_a$, we estimate the associated pixel as $\left[x+\Delta x(k_{max}), y+\Delta y(k_{max})\right]$.  
Otherwise there is no associated pixels in the neighborhood, and an occlusion is identified.

\section{Experimental Results}
\subsection{Comparison of Point-wise Full Velocity}

To the best of our knowledge, there is no existing method estimating {\it point-wise} full velocity for radar returns. 
Thus, we use point-wise radial velocity from raw radar returns as the baseline to compare with our estimation. 
We extract data from the nuScenes Object Detection Dataset~\cite{caesar2020nuscenes}, with $6432$, $632$, and $2041$ samples in training, validation and testing set, respectively. 
Each sample consists of a radar scan and two images for optical flow computation, {\it i.e.}, one image synchronizing with the radar and the other is a neighboring image frame. 
The optical flow is computed by the RAFT model~\cite{teed2020raft} pre-trained on KITTI~\cite{geiger2013vision}. 
The R2P network is an U-Net~\cite{ronneberger2015u, morris2018pyramid} with five levels of resolutions and $64$ channels for intermediate filters. 
The neighborhood skips every other pixel, and its size (in pixels) is (left: $4$, right: $4$, top: $10$, bottom: $4$) and an example of the neighborhood is illustrated in Fig.~\ref{fig:diagram_association}(b). 
The threshold of association scores $T_a$ is $0.3$. Parameters associating radar points with GT bounding box are set as $T_d=0.5$m and $T_p=20\%$. 
Parameter $c$ in Eq.~\eqref{eq:prob} is  $0.36$.
To obtain GT point-wise velocity, based on the criteria in Sec.~\ref{sec:GT_velocity}, we first associate moving radar points to GT detection boxes, whose GT velocity is assigned to associated points as their GT velocity. The GT velocity of bounding boxes is estimated from GT center positions in neighboring frames with timestamps. 

\begin{figure*}[t]
	\centering
	\captionsetup{font=small}
	\vspace{-2mm} 
	\scalebox{1.02}{
		\begin{tabular}{@{}c@{}c@{}c@{}c@{}}
			\includegraphics[height=1.09in]{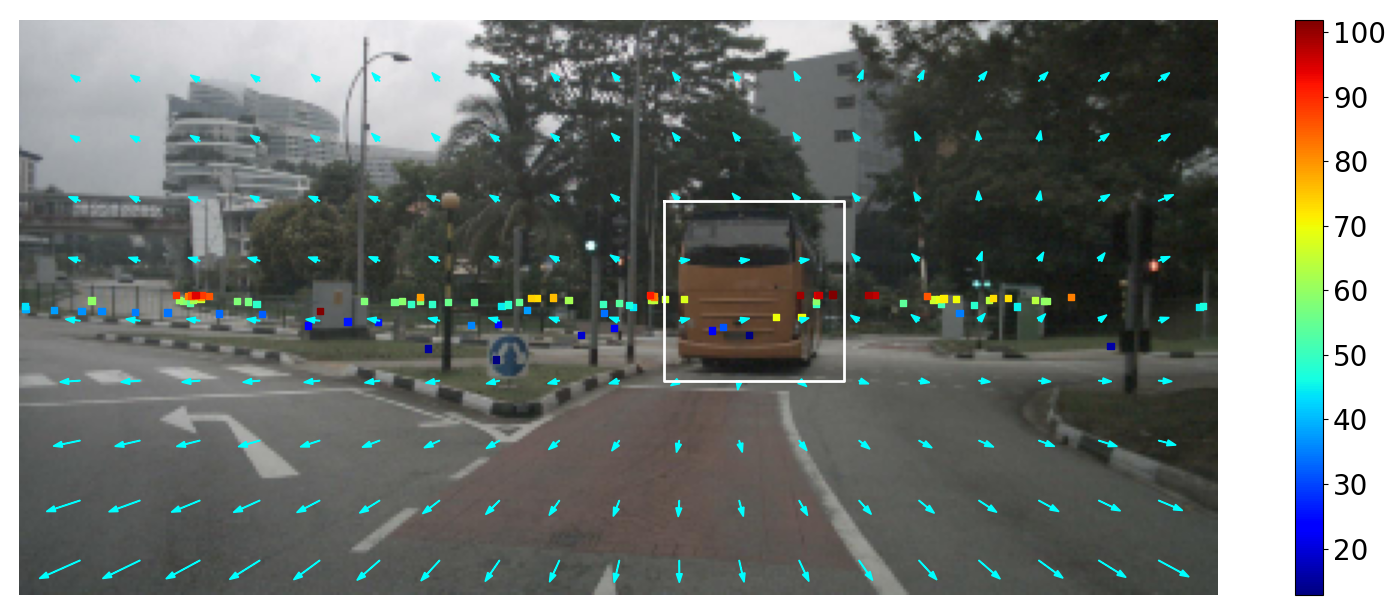}     &
			\includegraphics[height=1.16in]{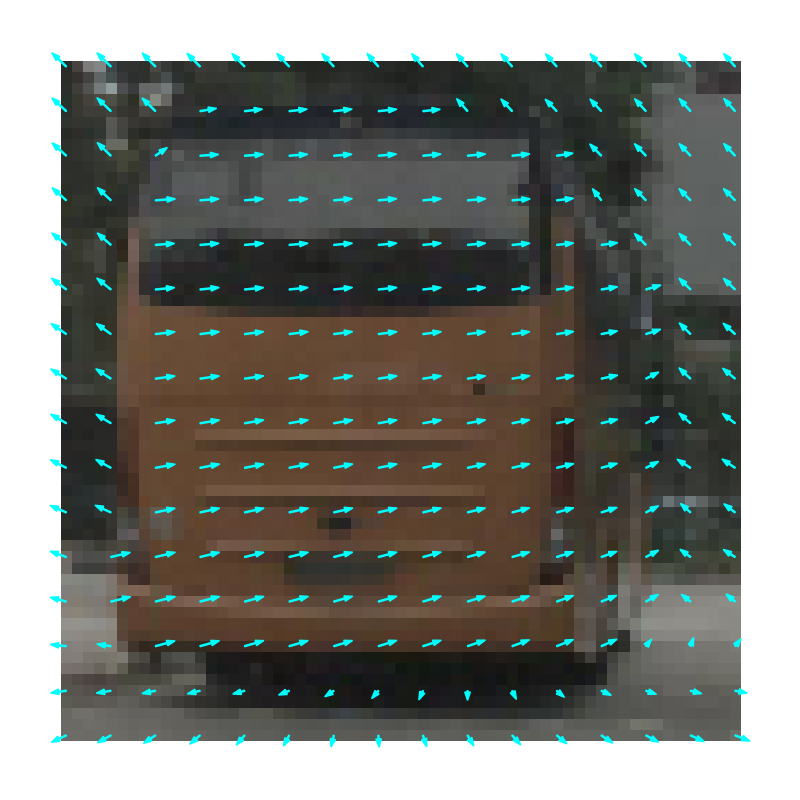} &
			\includegraphics[height=1.1in]{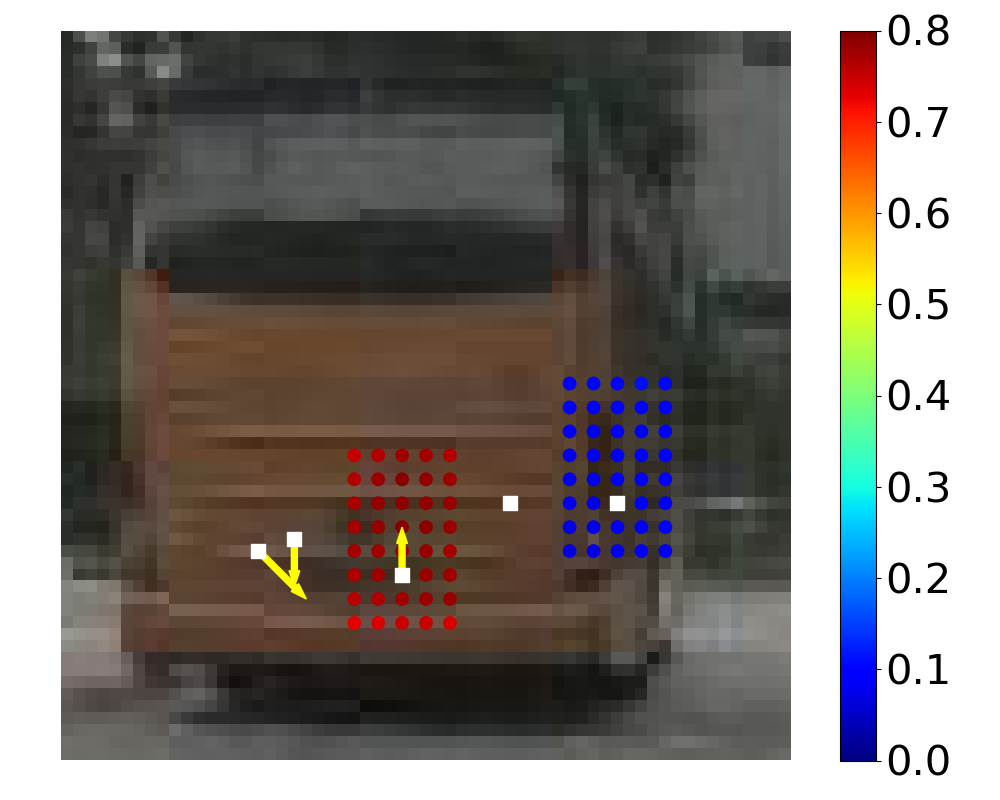}   &
			\includegraphics[height=1.1in]{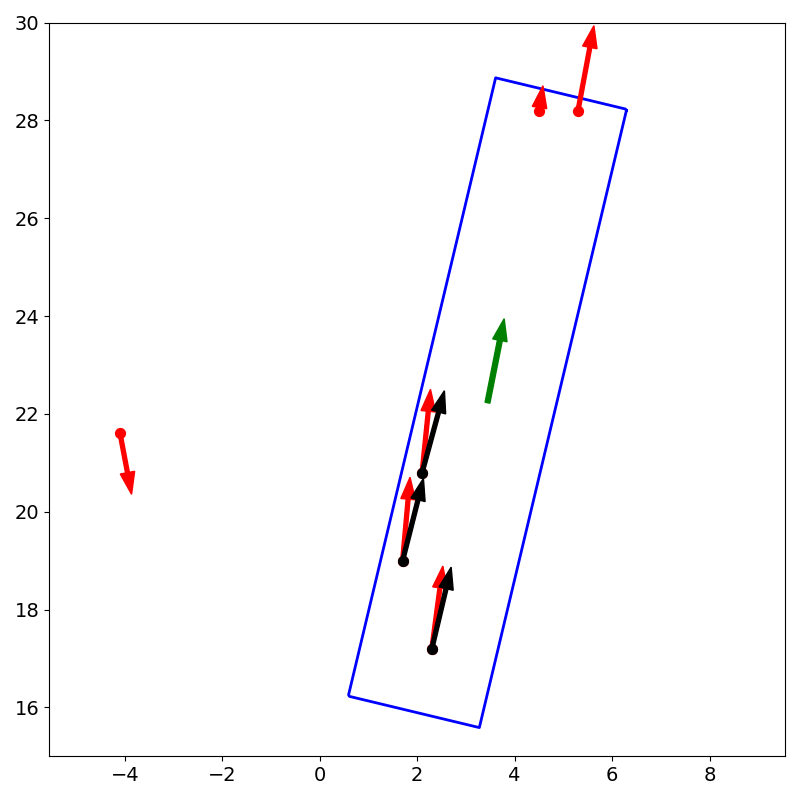}  \vspace{-2mm} \\
			\includegraphics[height=1.1in]{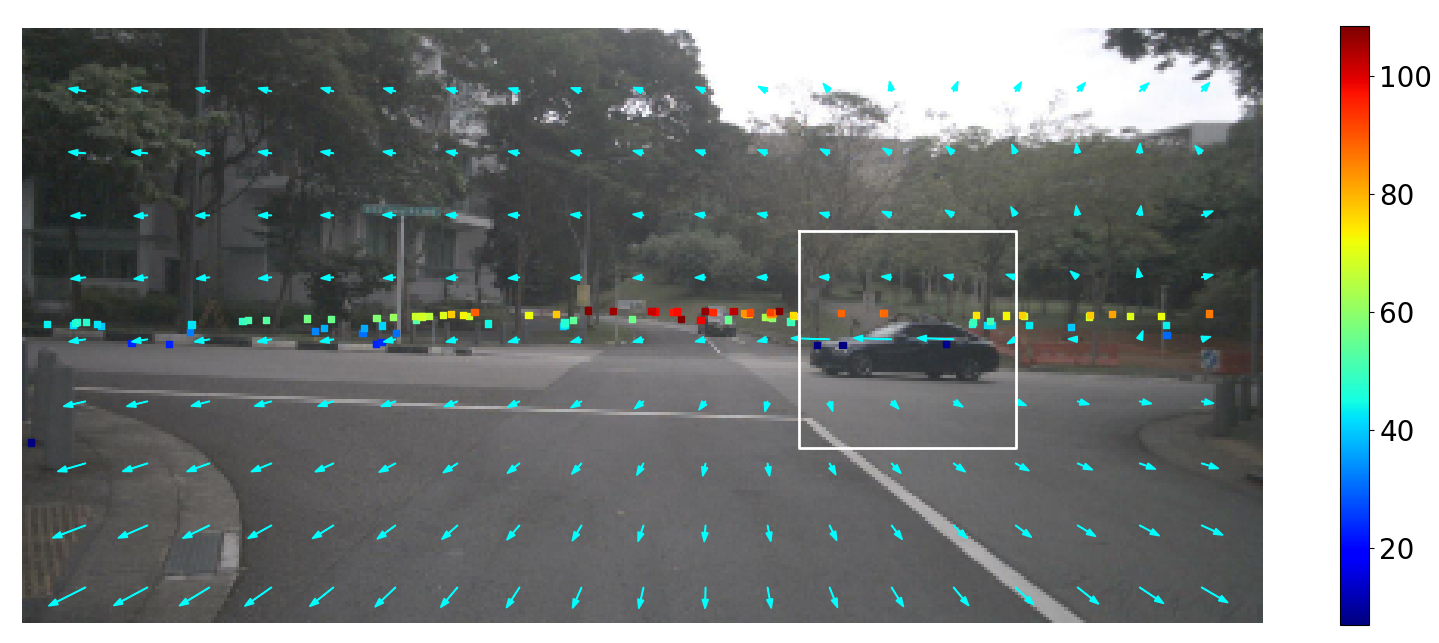}     &
			\includegraphics[height=1.12in]{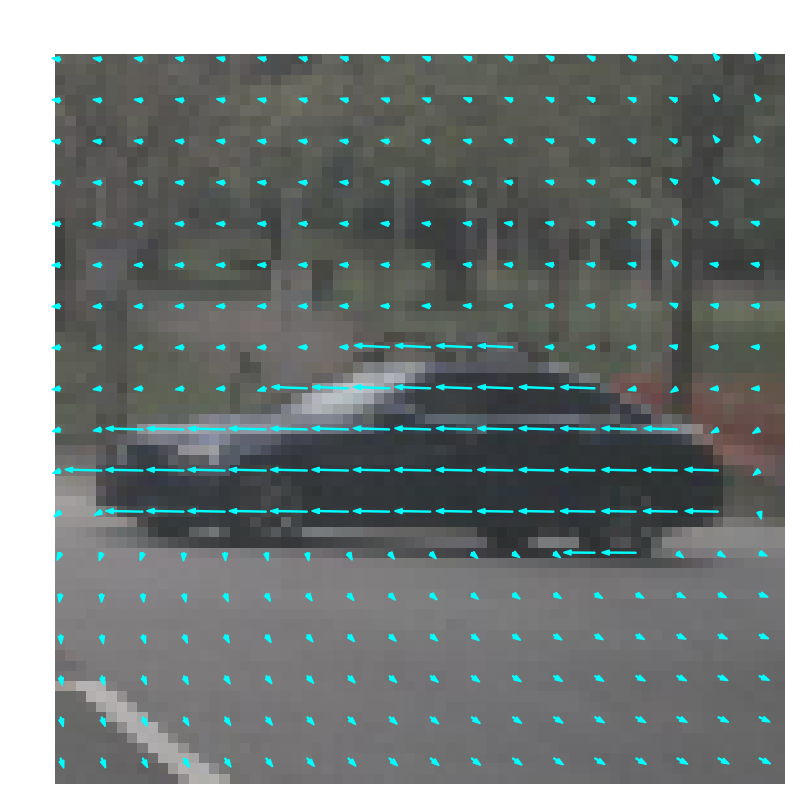} &
			\includegraphics[height=1.1in]{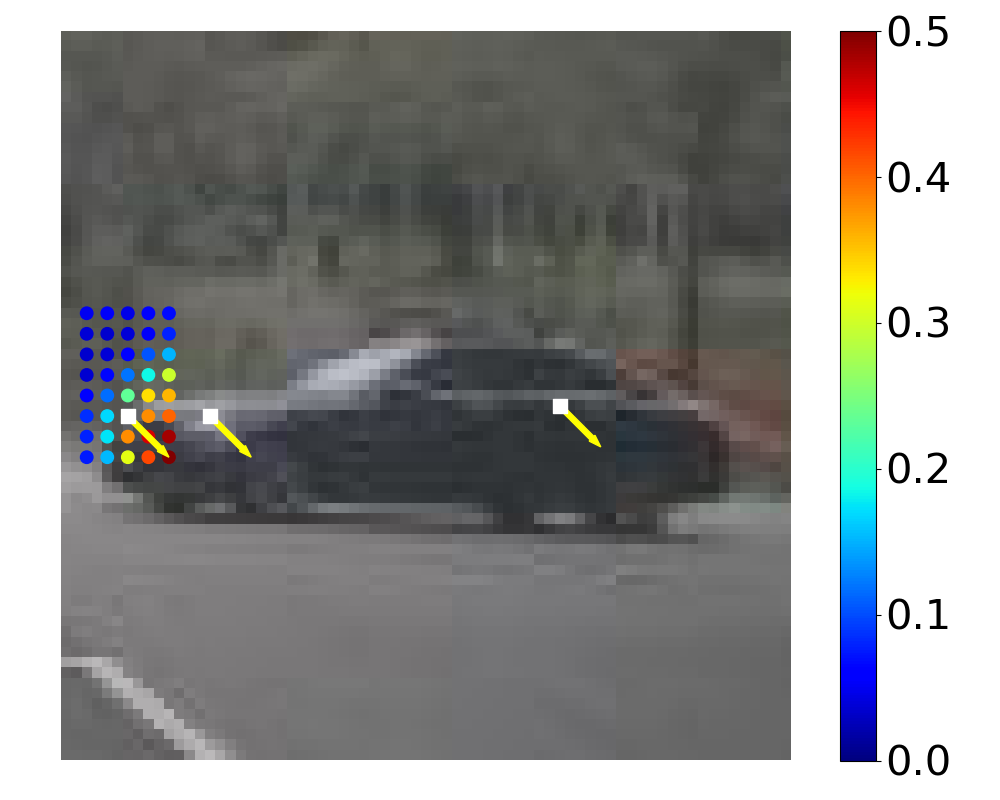}   &
			\includegraphics[height=1.1in]{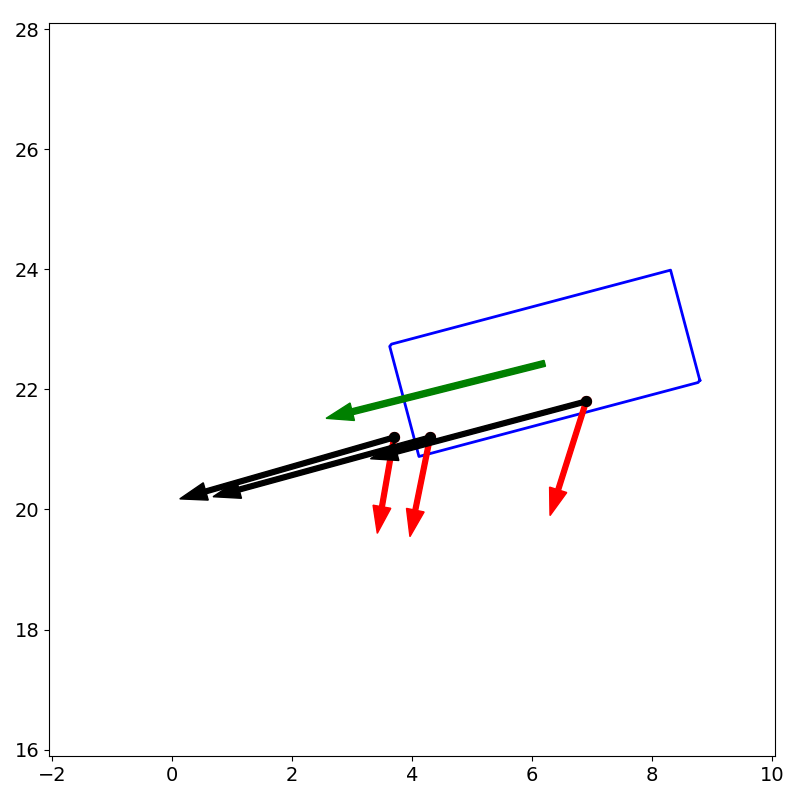}  \vspace{-3mm} \\
			\includegraphics[height=1.1in]{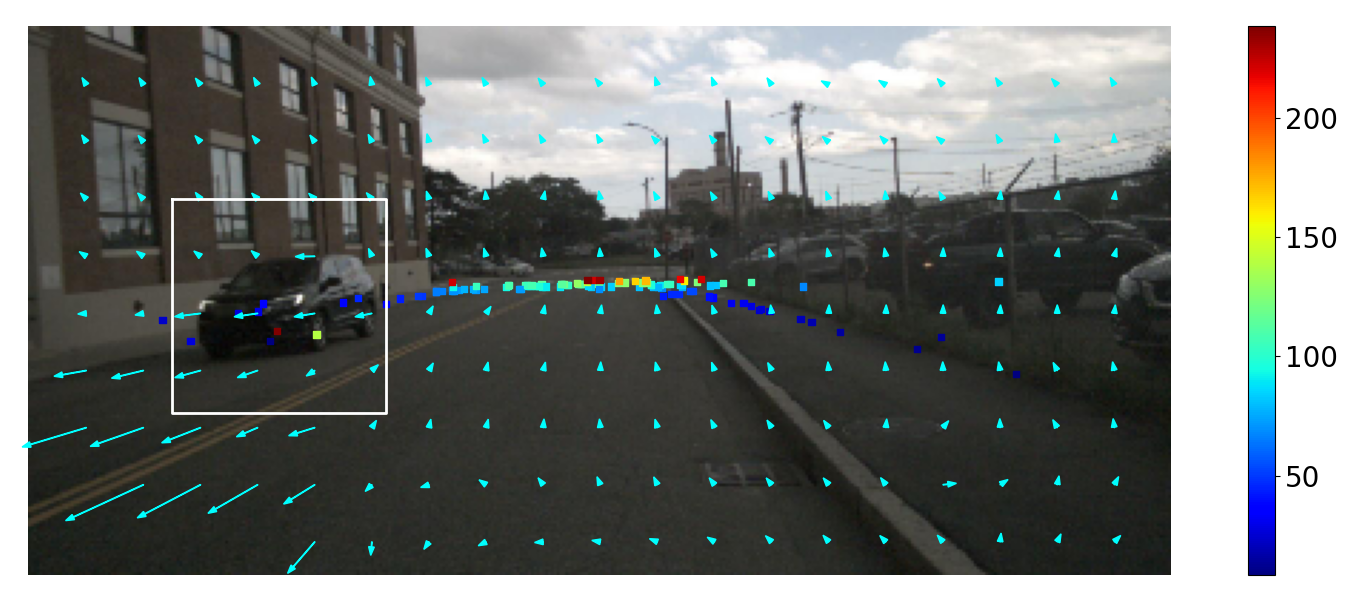}     &
			\includegraphics[height=1.15in]{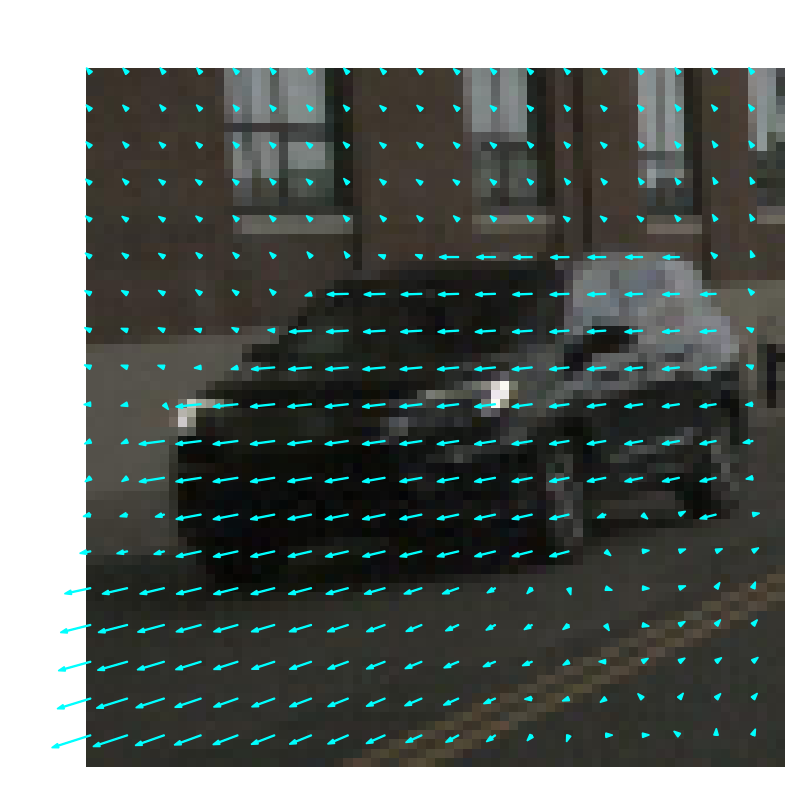} &
			\includegraphics[height=1.1in]{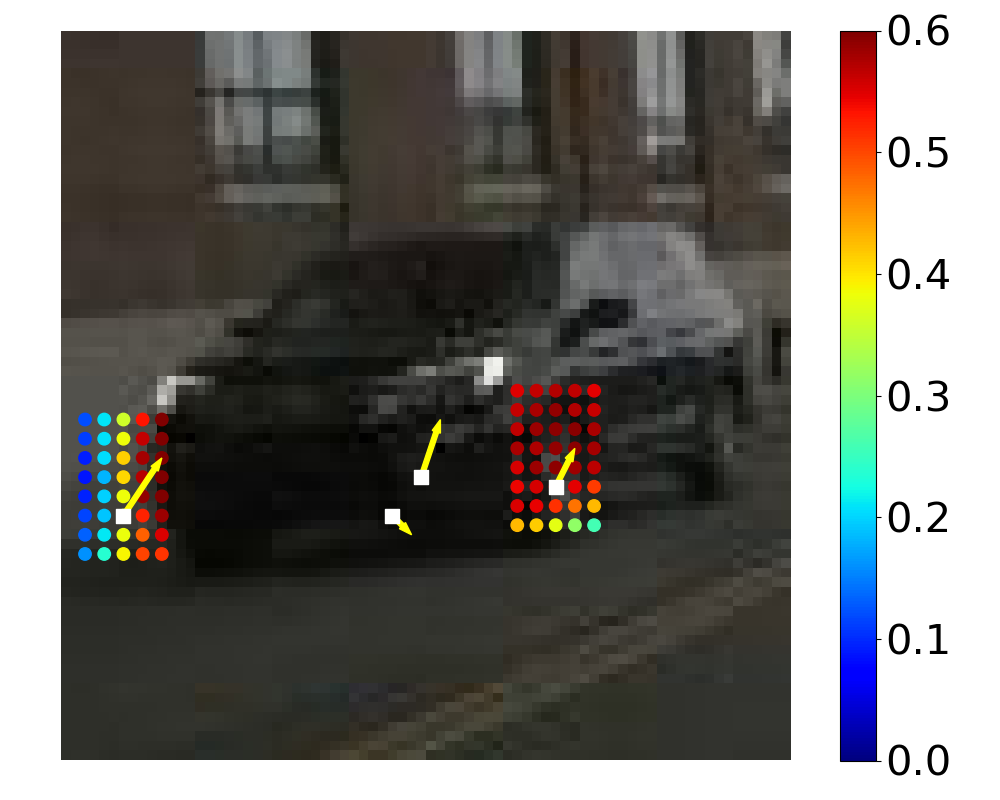}   &
			\includegraphics[height=1.1in]{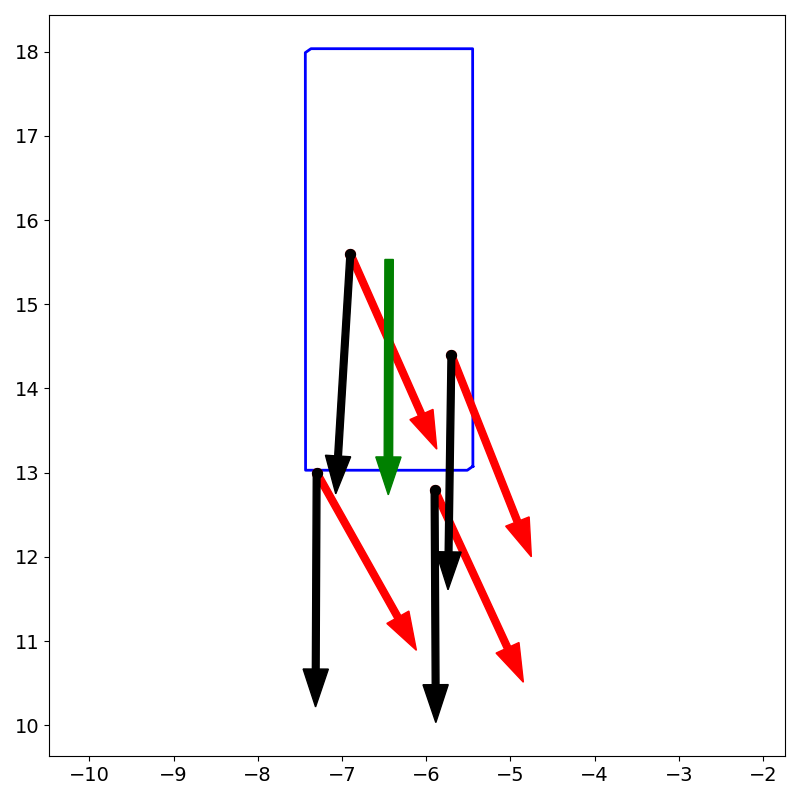}  \vspace{-2mm} \\
			\footnotesize{(a)} &  \footnotesize{ (b)} & \footnotesize{ (c)} & \footnotesize{ (d)} \vspace{-2mm}\\
	    \end{tabular} }
	\caption{\small Visualization of point-wise velocity estimation: (a) depth of all measured radar returns as well as flow, (b) optical flow in the white box region, (c) association scores around the selected radar projections as well as predicted mapping from raw radar projections to image pixels (yellow arrow) and (d) radial velocity (red), estimated full velocity (black) and GT velocity (green) in bird's-eye view.}
	\label{Figure:pv}
\end{figure*}

Tab.~\ref{tab:baseline} shows the average velocity error for moving points. The proposed method achieves substantially more accurate velocity estimation than the baseline. 
For instance, the error of our tangential component is only $21\%$ of that of the baseline.
We also have much smaller standard deviation, indicating more stable estimates. In addition, we list in Tab.~\ref{tab:baseline} velocity error of our method using raw radar projection for radar-camera association. Results show that, compared with using raw projection, using R2P network achieves higher estimation accuracy.
Fig.~\ref{Figure:pv} illustrates qualitative results of our point-wise velocity estimation.

\begin{table}[t!]
	\captionsetup{font=small}
	\begin{center}	
		\scalebox{0.75}{
			\begin{tabular}{|c|c|c|c|}
				\hline
				Mean Error (STD)  & Ours & Ours & Baseline \\ 
				(m/s) & (R2P Network) & (Raw Projection) & \\
				\hline
				Full Velocity  	      	& $\mathbf{0.433}\; (\mathbf{0.608})$ & $0.577\; (1.010)$ & $1.599\; (2.054)$ \\
				\hline
				Tangential Comp.    & $\mathbf{0.322}\; (\mathbf{0.610})$ & $0.472\; (1.024)$ & $1.536\; (2.083)$ \\
				\hline
				Radial Comp.    	& $0.205\; (0.196)$ & $0.205\; (0.196)$ & $0.205\; (0.196)$ \\
				\hline
			\end{tabular}
		}
	\end{center}
	\vspace{-5mm}
	\caption{\small Comparison of point-wise velocity error of our methods and the baseline (raw radial velocity).}
	\label{tab:baseline}
	\vspace{-2mm}
\end{table}

\subsection{Comparison of Object-wise Velocity}
Although there are no existing methods for point-wise velocity estimation for radar, a related work, CenterFusion~\cite{nabati2021centerfusion}, estimates {\it object-wise} full velocity  via object detection with image and radar inputs. 
To fairly compare with CenterFusion, we convert our point-wise velocity to object-wise velocity. 
Specifically, we use the average velocity of radar points associated with the same detected box as our estimate of object velocity. 
Points are associated with detected boxes according to distance. 
Note the point-wise velocity to object-wise velocity conversion is straightforward for comparison purposes, and there would be more advanced approaches to integrate point-wise full velocities in a detection network, which is beyond the scope of this work. 
Tab.~\ref{tab:det} shows that with our estimated full velocity, the velocity estimation for objects is significantly improved.

\begin{table}[t!]
	\captionsetup{font=small}
	\begin{center}	
		\scalebox{0.9}{
			\begin{tabular}{|c|c|}
				\hline
				Methods & Error (m/s) \\ 
				\hline
				Ours 	        & $\mathbf{0.451}$ \\
				CenterFusion~\cite{nabati2021centerfusion}    & $0.826$  \\
				\hline
			\end{tabular}
		}
	\end{center}
	\vspace{-5mm}
	\caption{\small Comparison of object-wise velocity errors. For a fair comparison we inherit the same set of detected objects from~\cite{nabati2021centerfusion}.}
	\label{tab:det}	\vspace{-3mm}
\end{table}

\subsection{Radar Point Accumulation}
Accumulating radar points over time can overcome the sparsity of radar hits acquired in a single sweep, achieving dense point cloud for objects and thus allowing techniques designed for processing LiDAR points to be applicable for radar. 
The point-wise velocity estimate makes it possible to compensate the motion of dynamic objects appearing in a temporal sequence of measurements for accumulation. 
Specifically, for a moving radar point (with estimated velocity $\mvdot$) in a previous frame $i$ captured at time $t_i$, its motion from $t_i$ to the time at the current frame, $t_0$, can be compensated by,
\begin{equation}
\bm{p_0}= \bm{p_i} + \mvdot(t_0 - t_i),
\label{eq:motion_compensate}
\end{equation}
where $\bm{p_i}$ and $\bm{p_0}$ are the radar point coordinates at $t_i$ and $t_0$ in radar coordinates of $t_i$. 
Then $\bm{p_0}$ is transformed to current radar coordinates by known egomotion from $t_i$ to $t_0$.

\begin{figure}[t!]
    \captionsetup{font=small}
	\begin{center}
		\includegraphics[width=0.8\linewidth]{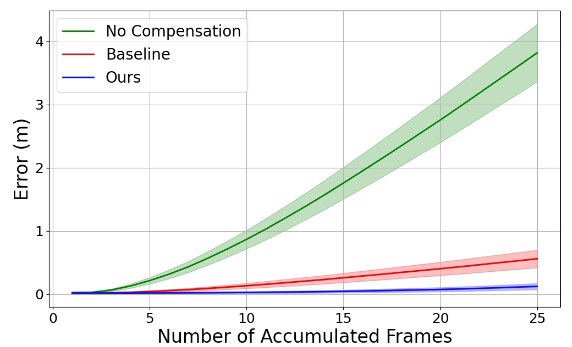}
	\end{center}
	\vspace{-5mm}
	\caption{\small Error comparison when accumulating radar points from increasing number of frames. The lines represent mean error and shaded area $\pm0.1\times$ STD. Our  full velocity based accumulation outperforms the ones with radial velocity, or no compensation.}
	\label{fig:curve}\vspace{-3mm}
\end{figure}

\begin{figure*}[t!]
	\centering
	\captionsetup{font=small}
	\scalebox{1}{
		\begin{tabular}{@{}c@{}c@{}c@{}c@{}c@{}}
			\includegraphics[width=1.2in]{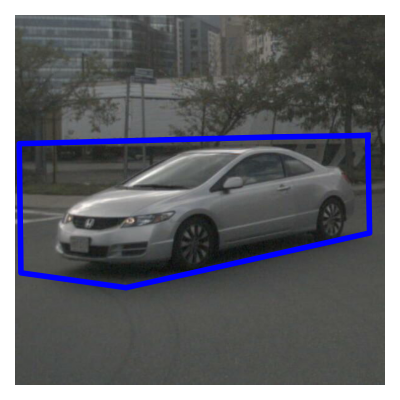} &
			\includegraphics[width=1.2in]{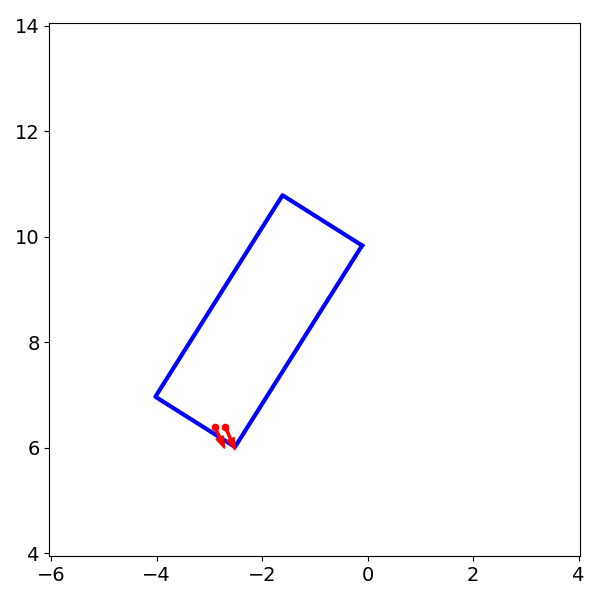}&
			\includegraphics[width=1.2in]{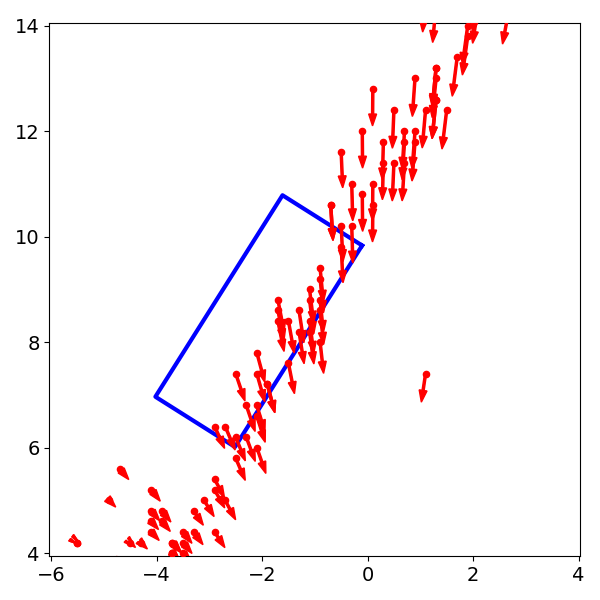} &
			\includegraphics[width=1.2in]{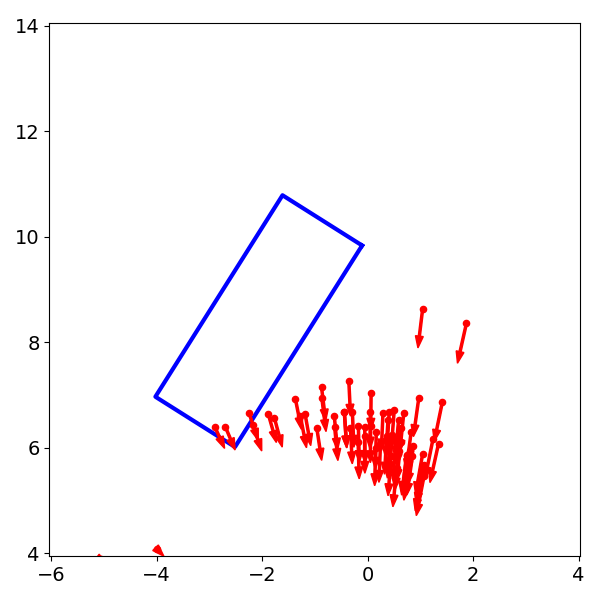}&
			\includegraphics[width=1.2in]{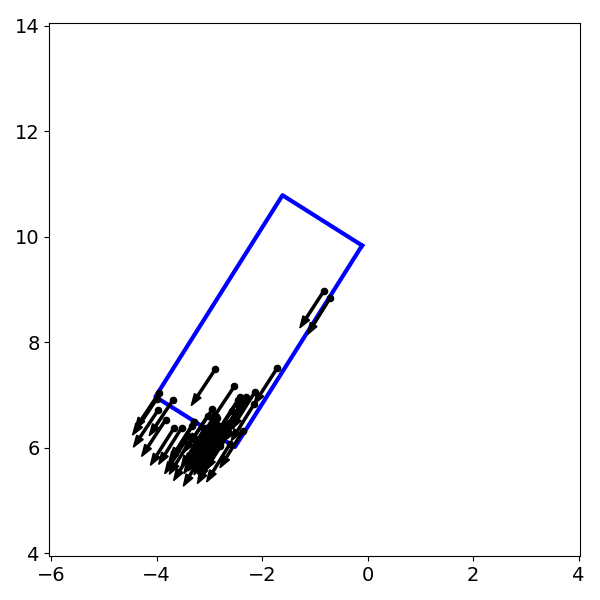}\vspace{-2mm}\\
			\includegraphics[width=1.2in]{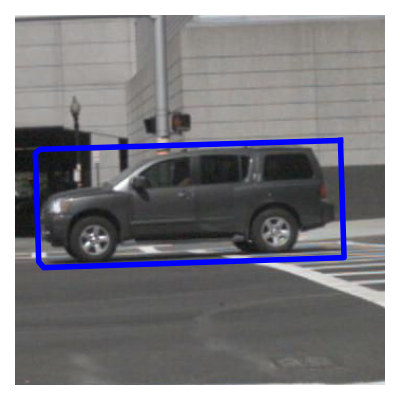} &
			\includegraphics[width=1.2in]{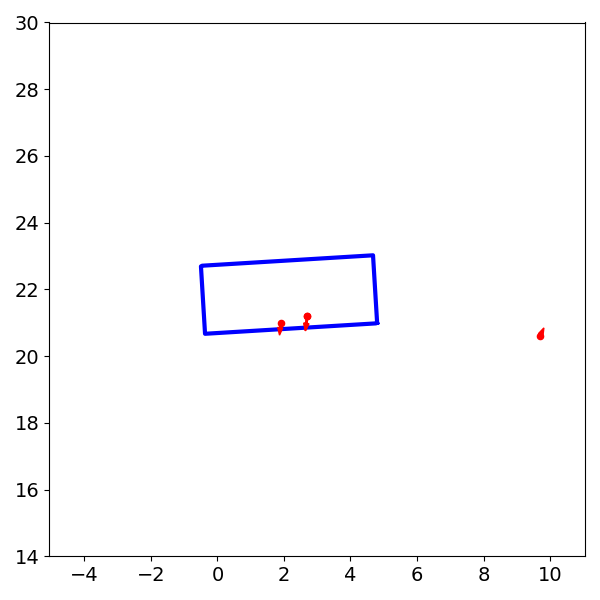}&
			\includegraphics[width=1.2in]{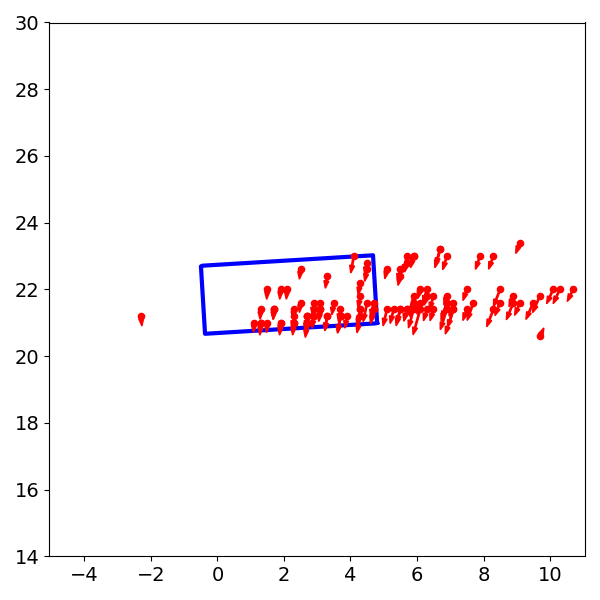} &
			\includegraphics[width=1.2in]{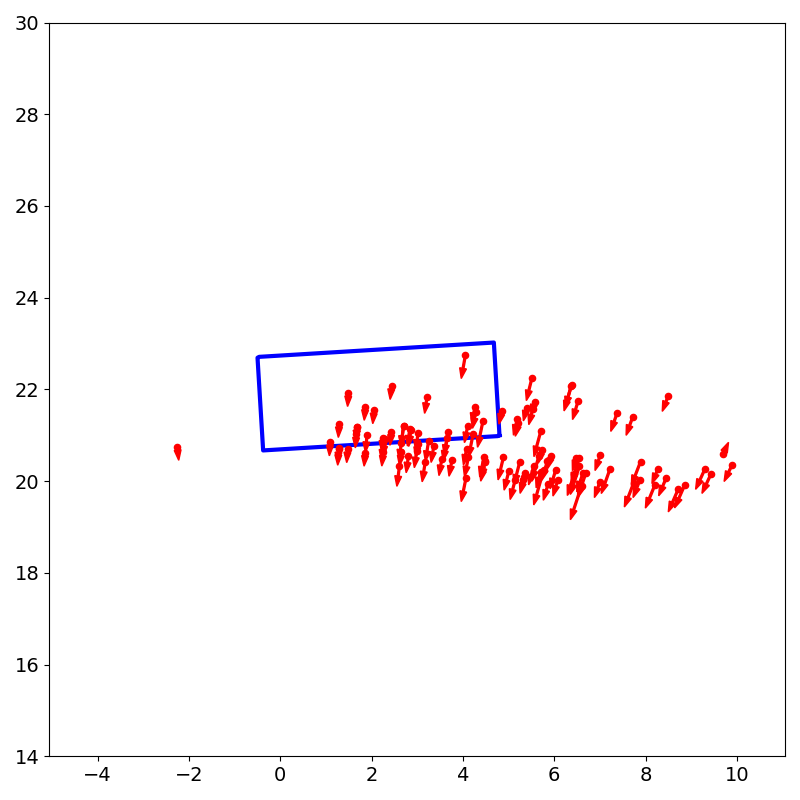}&
			\includegraphics[width=1.2in]{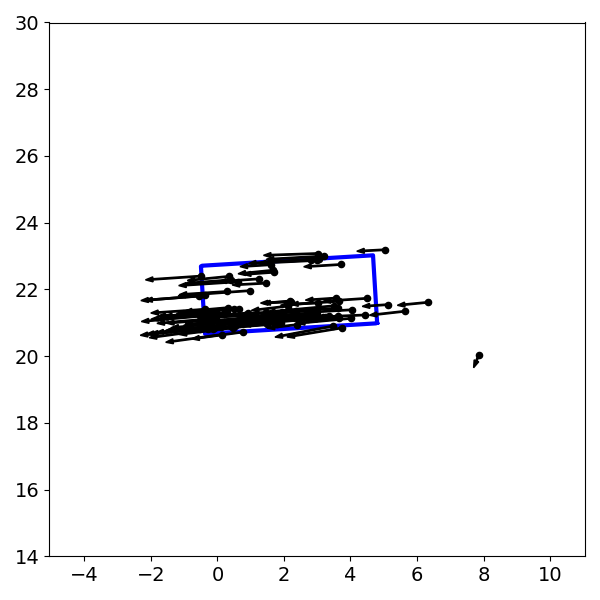}\vspace{-2mm}\\
			\includegraphics[width=1.2in]{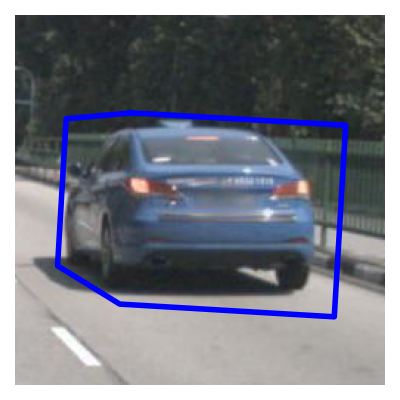} &
			\includegraphics[width=1.2in]{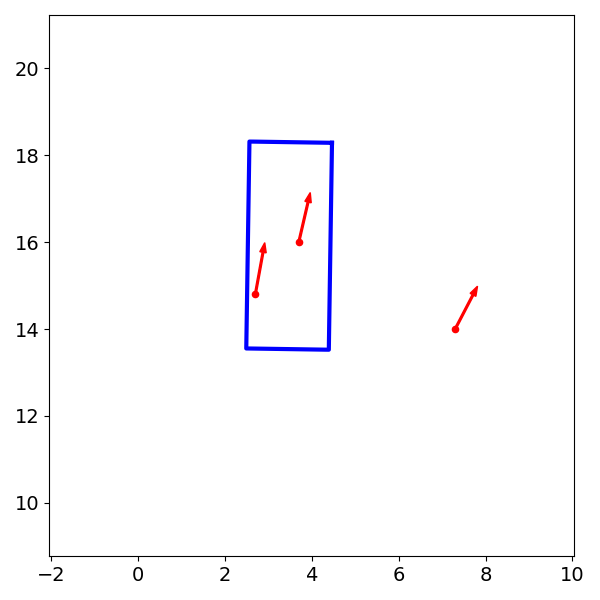}&
			\includegraphics[width=1.2in]{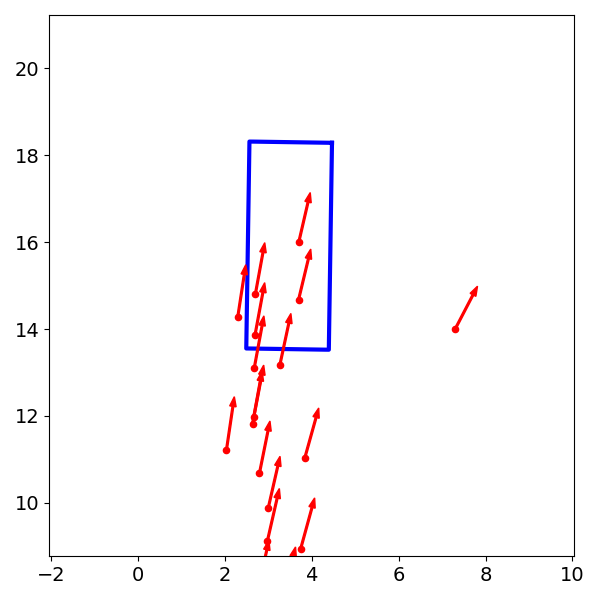} &
			\includegraphics[width=1.2in]{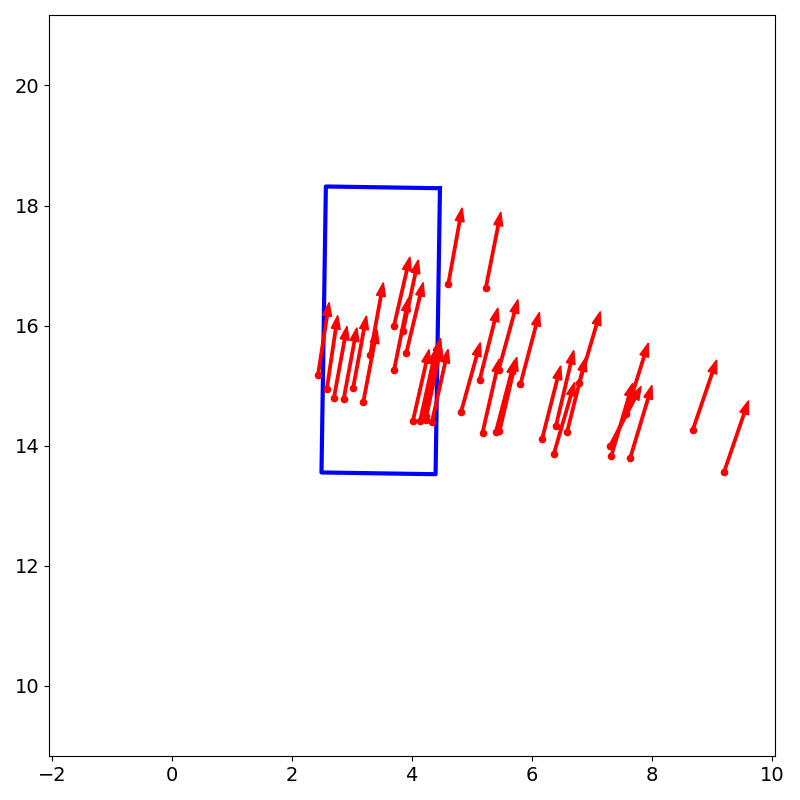}&
			\includegraphics[width=1.2in]{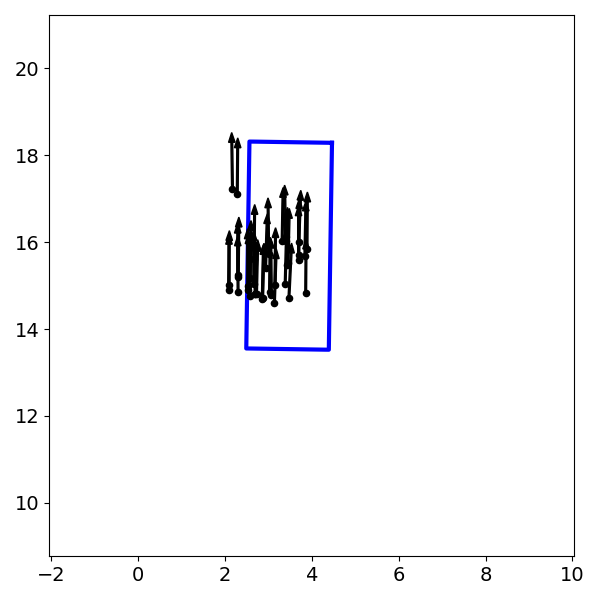}\vspace{-2mm}\\
			\includegraphics[width=1.2in]{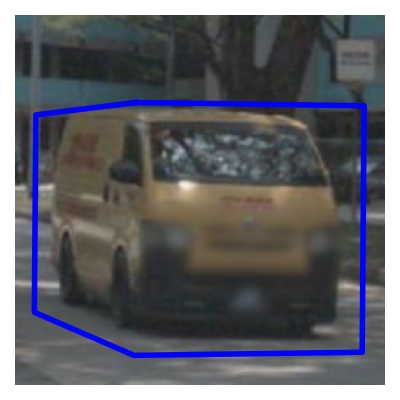} &
			\includegraphics[width=1.2in]{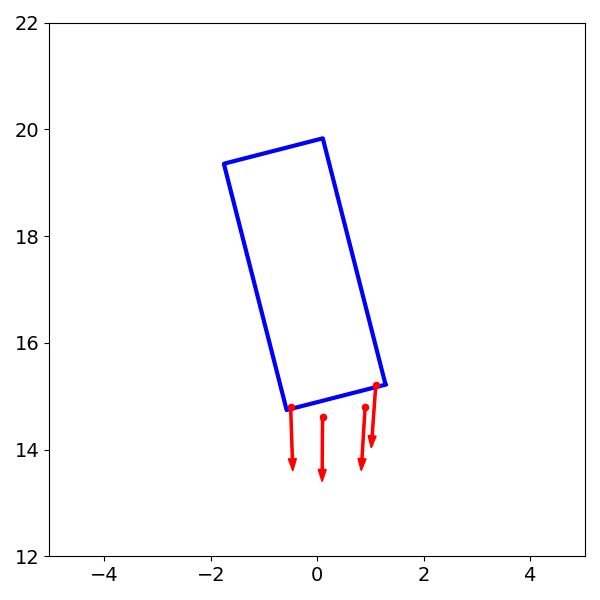}&
			\includegraphics[width=1.2in]{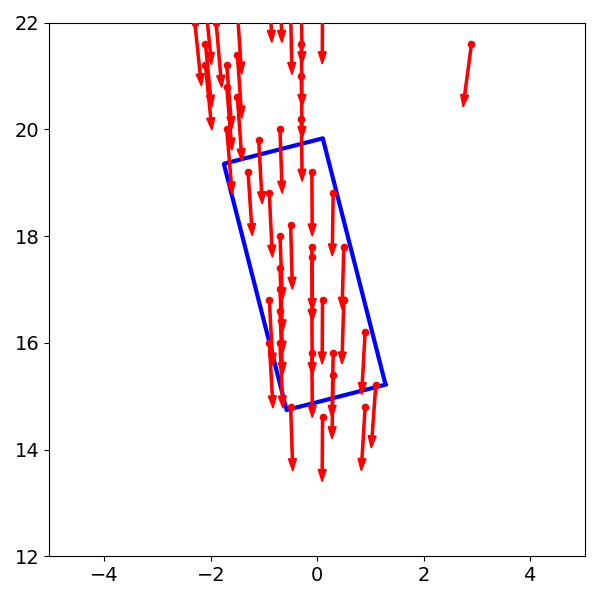} &
			\includegraphics[width=1.2in]{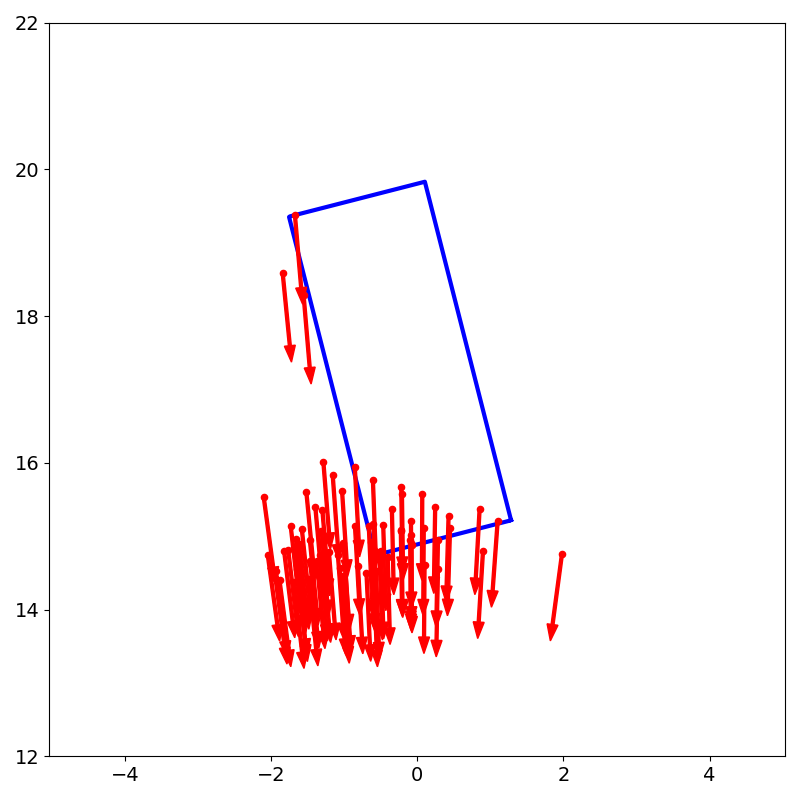}&
			\includegraphics[width=1.2in]{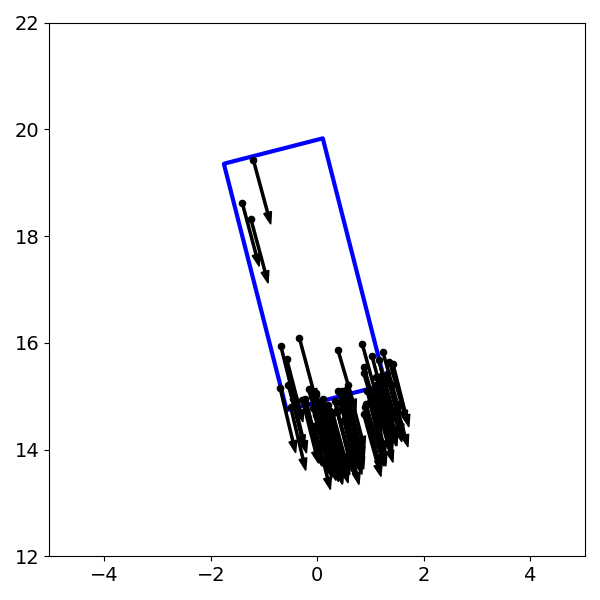}\vspace{-2mm}\\
			\footnotesize{(a)} &  \footnotesize{ (b)} & \footnotesize{ (c)} & \footnotesize{ (d)} & \footnotesize{ (e)}\\
	\end{tabular} }
	\vspace{-2mm}
	\caption{\small Moving radar points are plotted with point-wise radial (red) and full (black) velocity, including image with bounding box (a), single-frame radar points in bird's-eye view (b), accumulated radar points from $20$ frames without motion compensation (c), with radial velocity based compensation (d), and with our full-velocity based compensation (e).
	 Our accumulated points are tightly surrounding the bounding box, which will benefit downstream tasks such as pose estimation and object detection.}
	\label{Figure:acc}	\vspace{-1mm}
\end{figure*}

\Paragraph{Qualitative results}
Fig.~\ref{Figure:acc} shows accumulated points of moving vehicles in radar coordinates. 
For comparison, we show accumulated radar points compensated by our estimated full velocity, compensated with radial velocity (baseline) and without motion compensation. Compared with the baseline and no motion compensation, our accumulated points are more consistent with the GT bounding boxes.

\Paragraph{Quantitative results}
To quantitatively evaluate the accuracy of radar point accumulation, we use the mean distance from accumulated points (of up to $25$ frames) to their corresponding GT boxes as the accumulation error. 
This distance for points inside the box is zero, and outside it is the distance from the radar point to the closest point on the box's boundary. 
In Fig.~\ref{fig:curve}, we compare the accumulation for our method, the baseline and accumulation without motion compensation. 
While error increases with the number of frames for all methods, our method has the lowest rate of error escalation.

\begin{table}[t!]
    \captionsetup{font=small}
	\begin{center}	
		\scalebox{0.85}{
			\begin{tabular}{|c|c|c|}
				\hline
				  		                    Metric	& Ours &  Baseline \\ 
				\hline				
					Center Error (m) $\downarrow$            & $\mathbf{0.834}$ & $0.997$  \\			
					Orientation Error (degree) $\downarrow$  & $\mathbf{6.873}$ & $7.517$ \\
					IoU $\uparrow$ 						& $\mathbf{0.546}$ & $0.462$ \\
				\hline
			\end{tabular}
		}
	\end{center}
	\vspace{-4mm}
	\caption{\small Comparison of pose estimation performance: average error in center and orientation as well as Intersection over Union (IoU), by using BoxNet~\cite{nezhadarya2019boxnet} on radar points accumulated using our velocity and the radial velocity as a baseline.}
	\label{tab:iou}	\vspace{-3mm}
\end{table}

\Paragraph{Application of pose estimation}
To demonstrate the utility of accumulated radar points for downstream applications, we apply a pose estimation method, {\it i.e.}, BoxNet~\cite{nezhadarya2019boxnet}, on the accumulated 2D radar points via our full velocity and radial velocity (baseline), respectively. BoxNet takes pre-segmented 2D point clouds of an object as input and predicts a 2D bounding box with parameters as center position, length, width and orientation. 
We use accumulated radar points of $5702$, $559$ and $2001$ moving vehicles with corresponding GT bounding boxes as training, validation and testing data, respectively. Tab.~\ref{tab:iou} shows our accumulated radar achieves higher accuracy than the baseline.

\Section{Conclusion}
A drawback of Doppler radar has been that it provides only the radial component of velocity, which limits its utility in object velocity estimation, motion prediction and radar return accumulation. This paper addresses this drawback by presenting a closed-form solution to the full velocity of radar returns.  It leverages optical flow constraints to upgrade radial velocity into full velocity.  As part of this work, we use GT bounding-box velocities to supervise a  network that predicts association corrections for the raw radar projections.  We experimentally verify the effectiveness of our method and demonstrate its application on motion compensation for integrating radar sweeps over time.

This method developed here may apply to additional modalities such as full-velocity estimation from Doppler LiDAR and cameras.  

\vspace{1mm}
\noindent\textbf{Acknowledgement}
This work was supported by the Ford-MSU Alliance.

{\small
\bibliographystyle{ieee_fullname}
\bibliography{references}

\begin{thebibliography}{10}\itemsep=-1pt

\bibitem{barnes2020oxford}
Dan Barnes, Matthew Gadd, Paul Murcutt, Paul Newman, and Ingmar Posner.
\newblock The {Oxford Radar RobotCar Dataset}: A radar extension to the {Oxford
  RobotCar Dataset}.
\newblock In {\em IEEE International Conference on Robotics and Automation},
  pages 6433--6438, 2020.

\bibitem{m3d-rpn-monocular-3d-region-proposal-network-for-object-detection}
Garrick Brazil and Xiaoming Liu.
\newblock {M3D-RPN}: Monocular {3D} region proposal network for object
  detection.
\newblock In {\em IEEE International Conference on Computer Vision}, pages
  9287--9296, 2019.

\bibitem{kinematic-3d-object-detection-in-monocular-video}
Garrick Brazil, Gerard Pons-Moll, Xiaoming Liu, and Bernt Schiele.
\newblock Kinematic 3{D} object detection in monocular video.
\newblock In {\em European Conference on Computer Vision}, pages 135--152,
  2020.

\bibitem{brodeski2019deep}
Daniel Brodeski, Igal Bilik, and Raja Giryes.
\newblock Deep radar detector.
\newblock In {\em IEEE Radar Conference}, pages 1--6, 2019.

\bibitem{caesar2020nuscenes}
Holger Caesar, Varun Bankiti, Alex~H Lang, Sourabh Vora, Venice~Erin Liong,
  Qiang Xu, Anush Krishnan, Yu Pan, Giancarlo Baldan, and Oscar Beijbom.
\newblock {nuScenes}: A multimodal dataset for autonomous driving.
\newblock In {\em IEEE Conference on Computer Vision and Pattern Recognition},
  pages 11621--11631, 2020.

\bibitem{chadwick2019distant}
Simon Chadwick, Will Maddern, and Paul Newman.
\newblock Distant vehicle detection using radar and vision.
\newblock In {\em IEEE International Conference on Robotics and Automation},
  pages 8311--8317, 2019.

\bibitem{chang2020spatial}
Shuo Chang, Yifan Zhang, Fan Zhang, Xiaotong Zhao, Sai Huang, Zhiyong Feng, and
  Zhiqing Wei.
\newblock Spatial attention fusion for obstacle detection using {mmWave} radar
  and vision sensor.
\newblock {\em Sensors}, 20(4):956, 2020.

\bibitem{danzer20192d}
Andreas Danzer, Thomas Griebel, Martin Bach, and Klaus Dietmayer.
\newblock {2D} car detection in radar data with {PointNets}.
\newblock In {\em IEEE Intelligent Transportation Systems Conference}, pages
  61--66, 2019.

\bibitem{elfes1989using}
Alberto Elfes.
\newblock Using occupancy grids for mobile robot perception and navigation.
\newblock {\em Computer}, 22(6):46--57, 1989.

\bibitem{fritsche2017fusion}
Paul Fritsche, Bj{\"o}rn Zeise, Patrick Hemme, and Bernardo Wagner.
\newblock Fusion of radar, {LiDAR} and thermal information for hazard detection
  in low visibility environments.
\newblock In {\em IEEE International Symposium on Safety, Security and Rescue
  Robotics}, pages 96--101, 2017.

\bibitem{geiger2013vision}
Andreas Geiger, Philip Lenz, Christoph Stiller, and Raquel Urtasun.
\newblock Vision meets robotics: The {KITTI} dataset.
\newblock {\em The International Journal of Robotics Research},
  32(11):1231--1237, 2013.

\bibitem{depth-completion-with-twin-surface-extrapolation-at-occlusion-boundaries}
Saif Imran, Xiaoming Liu, and Daniel Morris.
\newblock Depth completion with twin surface extrapolation at occlusion
  boundaries.
\newblock In {\em IEEE Conference on Computer Vision and Pattern Recognition},
  pages 2583--2592, 2021.

\bibitem{depth-coefficients-for-depth-completion}
Saif Imran, Yunfei Long, Xiaoming Liu, and Daniel Morris.
\newblock Depth coefficients for depth completion.
\newblock In {\em IEEE Conference on Computer Vision and Pattern Recognition},
  pages 12438--12447, 2019.

\bibitem{kampffmeyer2018connnet}
Michael Kampffmeyer, Nanqing Dong, Xiaodan Liang, Yujia Zhang, and Eric~P Xing.
\newblock {ConnNet}: A long-range relation-aware pixel-connectivity network for
  salient segmentation.
\newblock {\em IEEE Transactions on Image Processing}, 28(5):2518--2529, 2018.

\bibitem{kaul2020rss}
Prannay Kaul, Daniele De~Martini, Matthew Gadd, and Paul Newman.
\newblock {RSS-Net}: Weakly-supervised multi-class semantic segmentation with
  {FMCW} radar.
\newblock In {\em IEEE Intelligent Vehicles Symposium}, pages 431--436, 2020.

\bibitem{kellner2013instantaneous}
Dominik Kellner, Michael Barjenbruch, Klaus Dietmayer, Jens Klappstein, and
  J{\"u}rgen Dickmann.
\newblock Instantaneous lateral velocity estimation of a vehicle using
  {Doppler} radar.
\newblock In {\em International Conference on Information Fusion}, pages
  877--884, 2013.

\bibitem{kellner2014instantaneous}
Dominik Kellner, Michael Barjenbruch, Jens Klappstein, J{\"u}rgen Dickmann, and
  Klaus Dietmayer.
\newblock Instantaneous full-motion estimation of arbitrary objects using dual
  {Doppler} radar.
\newblock In {\em IEEE Intelligent Vehicles Symposium}, pages 324--329, 2014.

\bibitem{li2020lidar}
You Li and Javier Ibanez-Guzman.
\newblock Lidar for autonomous driving: The principles, challenges, and trends
  for automotive lidar and perception systems.
\newblock {\em IEEE Signal Processing Magazine}, 37(4):50--61, 2020.

\bibitem{li2020deep}
Ying Li, Lingfei Ma, Zilong Zhong, Fei Liu, Michael~A Chapman, Dongpu Cao, and
  Jonathan Li.
\newblock Deep learning for {LiDAR} point clouds in autonomous driving: a
  review.
\newblock {\em IEEE Transactions on Neural Networks and Learning Systems},
  2020.

\bibitem{lim2019radar}
Teck-Yian Lim, Amin Ansari, Bence Major, Daniel Fontijne, Michael Hamilton,
  Radhika Gowaikar, and Sundar Subramanian.
\newblock Radar and camera early fusion for vehicle detection in advanced
  driver assistance systems.
\newblock In {\em Conference on Neural Information Processing Systems
  Workshops}, 2019.

\bibitem{lombacher2016potential}
Jakob Lombacher, Markus Hahn, J{\"u}rgen Dickmann, and Christian W{\"o}hler.
\newblock Potential of radar for static object classification using deep
  learning methods.
\newblock In {\em IEEE MTT-S International Conference on Microwaves for
  Intelligent Mobility}, pages 1--4, 2016.

\bibitem{long2021radar}
Yunfei Long, Daniel Morris, Xiaoming Liu, Marcos Castro, Punarjay Chakravarty,
  and Praveen Narayanan.
\newblock Radar-camera pixel depth association for depth completion.
\newblock In {\em IEEE Conference on Computer Vision and Pattern Recognition},
  pages 12507--12516, 2021.

\bibitem{morris2018pyramid}
Daniel Morris.
\newblock A pyramid {CNN} for dense-leaves segmentation.
\newblock In {\em Conference on Computer and Robot Vision}, pages 238--245.
  IEEE, 2018.

\bibitem{nabati2021centerfusion}
Ramin Nabati and Hairong Qi.
\newblock {CenterFusion}: Center-based radar and camera fusion for {3D} object
  detection.
\newblock In {\em IEEE Winter Conference on Applications of Computer Vision},
  pages 1527--1536, 2021.

\bibitem{nezhadarya2019boxnet}
Ehsan Nezhadarya, Yang Liu, and Bingbing Liu.
\newblock {BoxNet}: A deep learning method for {2D} bounding box estimation
  from bird's-eye view point cloud.
\newblock In {\em IEEE Intelligent Vehicles Symposium}, pages 1557--1564, 2019.

\bibitem{nobis2019deep}
Felix Nobis, Maximilian Geisslinger, Markus Weber, Johannes Betz, and Markus
  Lienkamp.
\newblock A deep learning-based radar and camera sensor fusion architecture for
  object detection.
\newblock In {\em Sensor Data Fusion: Trends, Solutions, Applications}, pages
  1--7. IEEE, 2019.

\bibitem{palffy2020cnn}
Andras Palffy, Jiaao Dong, Julian~FP Kooij, and Dariu~M Gavrila.
\newblock {CNN} based road user detection using the {3D} radar cube.
\newblock {\em IEEE Robotics and Automation Letters}, 5(2):1263--1270, 2020.

\bibitem{qi2017pointnet}
Charles~R Qi, Hao Su, Kaichun Mo, and Leonidas~J Guibas.
\newblock {PointNet}: Deep learning on point sets for {3D} classification and
  segmentation.
\newblock In {\em IEEE Conference on Computer Vision and Pattern Recognition},
  pages 652--660, 2017.

\bibitem{ronneberger2015u}
Olaf Ronneberger, Philipp Fischer, and Thomas Brox.
\newblock {U-Net}: Convolutional networks for biomedical image segmentation.
\newblock In {\em International Conference on Medical Image Computing and
  Computer-assisted Intervention}, pages 234--241, 2015.

\bibitem{roos2016reliable}
Fabian Roos, Dominik Kellner, J{\"u}rgen Dickmann, and Christian Waldschmidt.
\newblock Reliable orientation estimation of vehicles in high-resolution radar
  images.
\newblock {\em IEEE Transactions on Microwave Theory and Techniques},
  64(9):2986--2993, 2016.

\bibitem{scheiner2019multi}
Nicolas Scheiner, Nils Appenrodt, J{\"u}rgen Dickmann, and Bernhard Sick.
\newblock A multi-stage clustering framework for automotive radar data.
\newblock In {\em IEEE Intelligent Transportation Systems Conference}, pages
  2060--2067, 2019.

\bibitem{schlichenmaier2019clustering}
Johannes Schlichenmaier, Fabian Roos, Philipp H{\"u}gler, and Christian
  Waldschmidt.
\newblock Clustering of closely adjacent extended objects in radar images using
  velocity profile analysis.
\newblock In {\em IEEE MTT-S International Conference on Microwaves for
  Intelligent Mobility}, pages 1--4, 2019.

\bibitem{seyfiouglu2018deep}
Mehmet~Sayg{\i}n Seyfio{\u{g}}lu, Ahmet~Murat {\"O}zbayo{\u{g}}lu, and
  Sevgi~Zubeyde G{\"u}rb{\"u}z.
\newblock Deep convolutional autoencoder for radar-based classification of
  similar aided and unaided human activities.
\newblock {\em IEEE Transactions on Aerospace and Electronic Systems},
  54(4):1709--1723, 2018.

\bibitem{stanislas2015characterisation}
Leo Stanislas and Thierry Peynot.
\newblock Characterisation of the {Delphi} electronically scanning radar for
  robotics applications.
\newblock In {\em Australasian Conference on Robotics and Automation}, pages
  1--10, 2015.

\bibitem{teed2020raft}
Zachary Teed and Jia Deng.
\newblock {RAFT}: Recurrent all-pairs field transforms for optical flow.
\newblock In {\em European Conference on Computer Vision}, pages 402--419,
  2020.

\bibitem{wu2020deep}
Yutian Wu, Yueyu Wang, Shuwei Zhang, and Harutoshi Ogai.
\newblock Deep {3D} object detection networks using {LiDAR} data: A review.
\newblock {\em IEEE Sensors Journal}, 21(2):1152--1171, 2020.

\bibitem{yang2020radarnet}
Bin Yang, Runsheng Guo, Ming Liang, Sergio Casas, and Raquel Urtasun.
\newblock {RadarNet}: Exploiting radar for robust perception of dynamic
  objects.
\newblock In {\em European Conference on Computer Vision}, pages 496--512,
  2020.

\bibitem{zhao2019object}
Zhong-Qiu Zhao, Peng Zheng, Shou-tao Xu, and Xindong Wu.
\newblock Object detection with deep learning: A review.
\newblock {\em IEEE Transactions on Neural Networks and Learning Systems},
  30(11):3212--3232, 2019.

\end{thebibliography}
}

\title{\textbf{Full-Velocity Radar Returns by Radar-Camera Fusion \\
	-- Supplementary Material --}}
\author{}

\maketitle

\setcounter{figure}{0}
\setcounter{table}{0}
\setcounter{section}{0}

\begin{abstract}
	In the supplementary material, we illustrate predicted radar-camera association. Second, we evaluate the influences of two factors, depth and the angle between actual velocity and radial direction, on the performance of point-wise full velocity estimation. Moreover, we report the computational time of three components of the estimation pipeline. Finally, we present a video showing point-wise velocity estimation in real driving scenes.
\end{abstract}


\section{Visualization of Predicted Radar-Camera Association}
\begin{figure}[h]
	\begin{center}
		\includegraphics[width=0.4\linewidth]{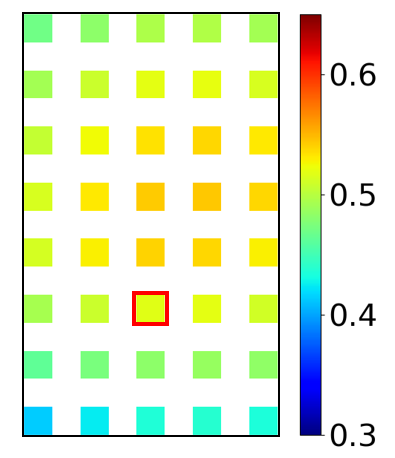}
	\end{center}
	\vspace{-5mm}
	\caption{\small Mean of predicted association over the test set in the neighborhood of raw projection (marked by a red square). The neighborhood region has a size of $9\times15$ pixels and association estimation skips every other pixel to improve the computational efficiency.}
	\label{fig:offset_prob}\vspace{-6mm}
\end{figure}

Fig.~\ref{fig:offset_prob} shows the mean of predicted association $A$ for the test set. 
It appears the radar point is more likely associated with pixels above the raw projection, as the measured radar height is always on the radar plane which is typically lower than vehicle height.

\section{Velocity Estimation Error for Different Depths and $\mathbf{\alpha}$}
\begin{figure*}[t]
	\centering
	\vspace{-2mm} 
	\scalebox{1.02}{
		\begin{tabular}{@{}c@{}c@{}c@{}}
			\includegraphics[width=2.2in]{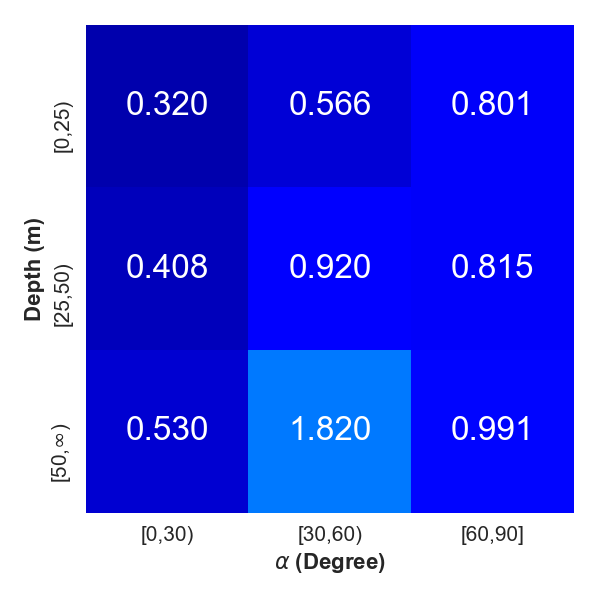} &
			\includegraphics[width=2.2in]{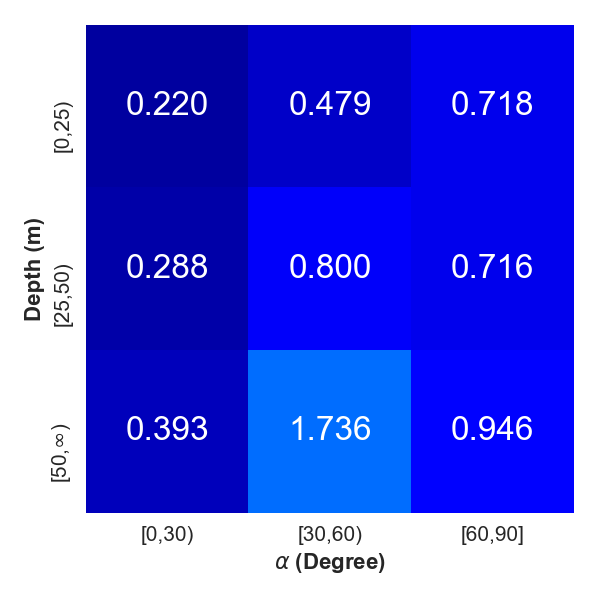} &
			\includegraphics[width=2.2in]{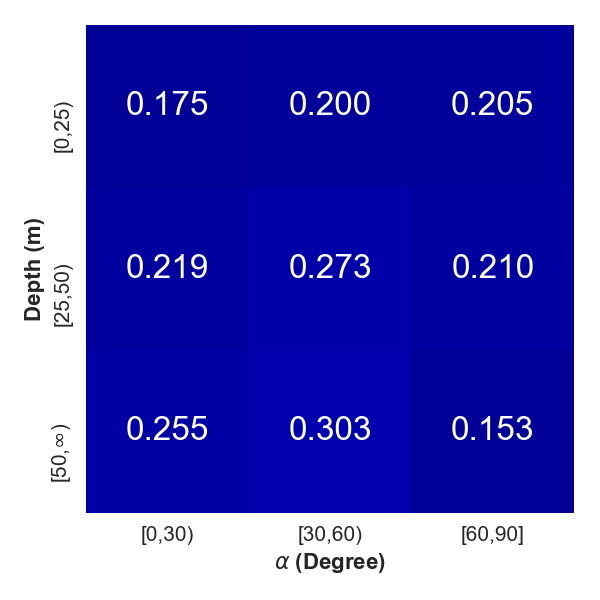}    \\
			\vspace{-4mm}\\
			\includegraphics[width=2.2in]{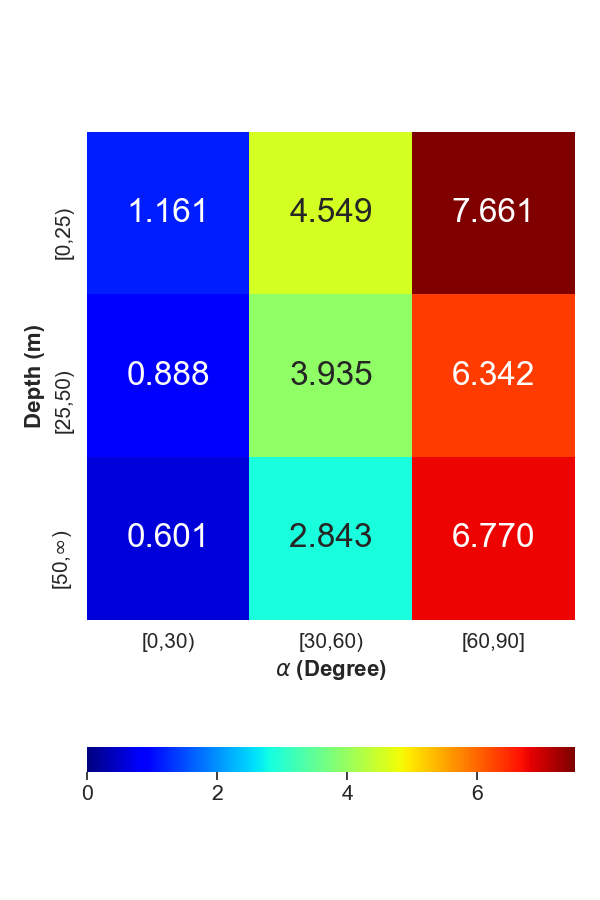}  &
			\includegraphics[width=2.2in]{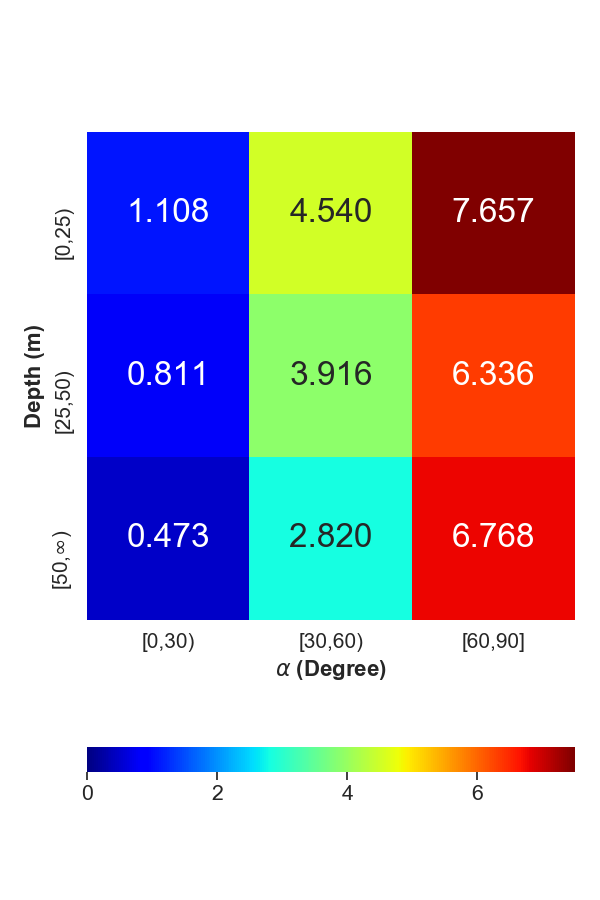} &
			\includegraphics[width=2.2in]{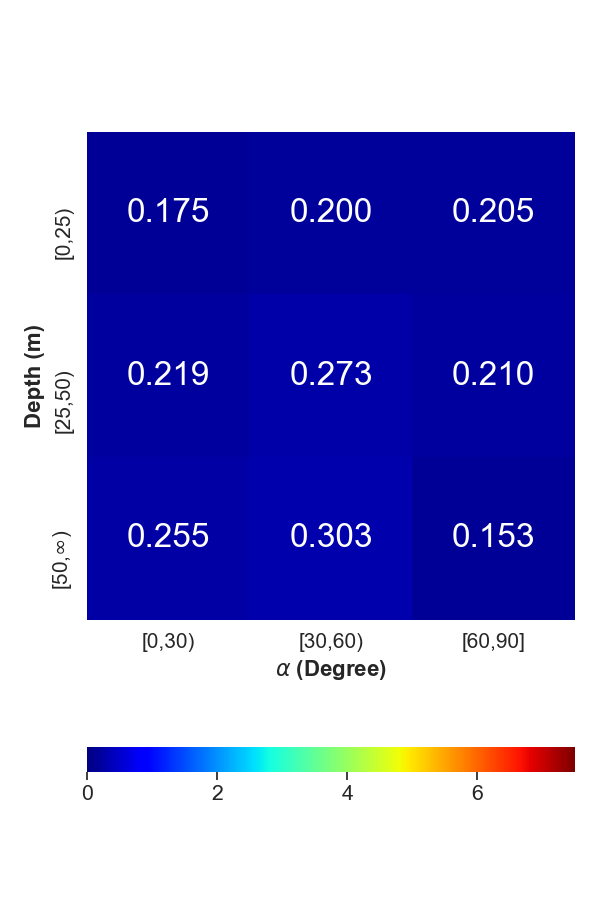}     \\
			\footnotesize{(a) Full Velocity Error} &  \footnotesize{ (b) Tangential Component Error} & \footnotesize{ (c) Radial Component Error}  \vspace{-2mm}\\
	\end{tabular} }
	\caption{\small Comparison of average error (in meters) of point-wise velocity estimates by the proposed method (first row) and baseline (second row). Columns 1, 2 and 3 are error of full velocity, tangential component and radial component, respectively. Each heat map shows the error for radar points in different depth and $\alpha$ ranges, where $\alpha$ is the angle between full velocity and radial direction.}
	\label{Figure:heatmap}
\end{figure*}

This experiment extends the evaluation of point-wise velocity estimation discussed in Section~4.1 of the main paper. In Fig.~\ref{Figure:heatmap}, each heat map shows point-wise velocity error under different depth ranges, i.e $[0,25)$, $[25,50)$ and $[50,\infty)$ meters as well as various $\alpha$ ranges, i.e. $[0,30)$, $[30,60)$ and $[60,90]$ degrees, where $\alpha$ is the angle between actual moving direction and radial direction of a radar point and ranges from 0 to 90 degrees. Results of the proposed method and baseline are show in the first row and second row, respectively.
The baseline (second row), with only radial measurement, suffers from large $\alpha$ since the the actual moving direction is very different from radial direction under large $\alpha$. The proposed method outperforms the baseline in all depth and $\alpha$ ranges for full velocity estimation. 

\section{Inference Time}
The pipeline of our full velocity estimation includes three major components, optical flow computation, radar-camera association estimation and closed-form solution of full velocity. The time used by each component per frame is listed in Table~\ref{tab:time}. Our computational platform includes Intel Core i7-8700 CPUs and a NVIDIA GeForce RTX 2080 Ti GPU. The proposed closed-form solution achieves highly efficient computation. Note the computational cost of optical flow can be improved by limiting the region of flow computation to areas with radar projections.

\begin{table}[h]
	\begin{center}	
		\begin{tabular}{|c|c|}
			\hline
			Components  				& Time per Frame (s)  \\ 
			\hline
			Optical Flow~\cite{teed2020raft} 				& $4.47\times 10^{-1}$ \\
			\hline
			Radar-camera Association    & $2.58\times 10^{-3}$ \\
			\hline
			Closed-form Velocity Computation   	& $2.49\times 10^{-4}$ \\
			\hline
		\end{tabular}
	\end{center}
	\vspace{-3mm}
	\caption{\small Computational time of three components in the method pipeline.}
	\label{tab:time}
\end{table}

\section{Video File}
In the video, we show point-wise velocity estimation (black arrow) of dynamic radar points in bird-eye view of radar coordinates. Moving radar points are also plotted with radial velocity (red arrow) and static points are shown in orange. The true velocity of vehicles are plotted as green arrow. The GT moving and static vehicles are plotted as solid and dashed bounding boxes, respectively. Images with radar projections are shown at top-left corner for reference.

\end{document}